\documentclass[11pt]{article}

\usepackage{times}
\usepackage[margin=1in]{geometry}
\usepackage[numbers]{natbib}
\usepackage[T1]{fontenc}
\usepackage{xcolor}
\usepackage{multicol}
\usepackage{enumitem}
\definecolor{mycitecolor}{RGB}{71, 191, 38}
\definecolor{mylinkcolor}{RGB}{40, 115, 201}

\usepackage[bookmarks=true, colorlinks, citecolor=mycitecolor,linkcolor=mylinkcolor,urlcolor=mycitecolor ]{hyperref}
\usepackage{mathtools} % load AMS maths
\usepackage{amsmath} % for math spacing
\usepackage{amssymb} % for math spacing
\usepackage{fancyhdr, bbm}
\usepackage{amsthm}
\usepackage{diagbox}
\usepackage{booktabs}
\usepackage{arydshln}
\usepackage{tikz}
\usetikzlibrary{patterns,angles,quotes,arrows}
\usepackage{subcaption}
\usepackage{algorithm}
\usepackage{algpseudocode}
\usepackage[normalem]{ulem}  % JUST TO ADD STRIKEOUTS

\newtheoremstyle{mystyle}%    % Name
  {}%                         % Space above
  {}%                         % Space below
  {}%                         % Body font
  {}%                         % Indent amount
  {\bfseries}%                % Theorem head font
  {.}%                        % Punctuation after theorem head
  { }%                        % Space after theorem head, ' ', or \newline
  {}%                         % Theorem head spec (can be left empty, meaning `normal')
\newtheorem{remark}{Remark}
\newtheorem{theorem}{Theorem}
\newtheorem{proposition}{Proposition}
\newtheorem{lemma}{Lemma}
\newtheorem{definition}{Definition}
\newtheorem{problem}{Problem}
\newtheorem{cor}{Corollary}

\usepackage[absolute]{textpos}

\begin{document}

\begin{textblock}{5}(0.5,0.5)
\noindent \textbf{Robotics: Science and Systems 2024} \\
\textbf{Delft, Netherlands, July 15-July 19, 2024}
\end{textblock}

\pagenumbering{arabic}
% Shrink space above figure
% \setlength{\intextsep}{5pt}
% % Shrink space between figure and caption
% \setlength{\abovecaptionskip}{2pt}
% % Shrink space after figure captions
% \setlength{\belowcaptionskip}{-2pt}

\title{POLICEd RL: Learning Closed-Loop Robot Control Policies with Provable Satisfaction of Hard Constraints}

\author{Jean-Baptiste Bouvier, Kartik Nagpal and Negar Mehr \\
\href{https://iconlab.negarmehr.com/}{ICON Lab}, Department of Mechanical Engineering\\
University of California Berkeley\\
Email: \{bouvier3, kartiknagpal, negar\}@berkeley.edu}

\date{}

\maketitle

\begin{abstract}
    In this paper, we seek to learn a robot policy guaranteed to satisfy state constraints. To encourage constraint satisfaction, existing RL algorithms typically rely on Constrained Markov Decision Processes and discourage constraint violations through reward shaping. However, such \emph{soft constraints} cannot offer verifiable safety guarantees. To address this gap, we propose \emph{POLICEd RL}, a novel RL algorithm explicitly designed to enforce affine \emph{hard constraints} in closed-loop with a black-box environment. Our key insight is to force the learned policy to be affine around the unsafe set and use this affine region as a repulsive buffer to prevent trajectories from violating the constraint. We prove that such policies exist and guarantee constraint satisfaction. Our proposed framework is applicable to both continuous and discrete state-action spaces and is agnostic to the choice of the RL training algorithm. Our results demonstrate the capacity of POLICEd RL to enforce hard constraints in robotic tasks while significantly outperforming existing methods. Code available at \href{https://iconlab.negarmehr.com/POLICEd-RL/}{https://iconlab.negarmehr.com/POLICEd-RL/}   
\end{abstract}

\section{Introduction}

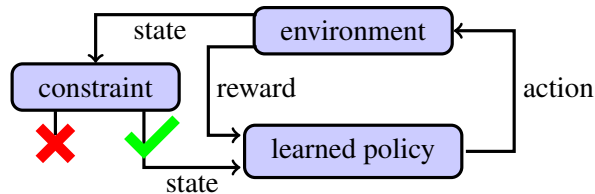
\begin{figure}[htbp!]
    \centering
    \tikzstyle{mybox} = [draw=black, fill=blue!20, very thick, rectangle, rounded corners, inner sep=10pt, inner ysep=5pt]
    \begin{tikzpicture}[scale = 1.07]
        \node[mybox] at (0, 0) {learned policy};
        \draw[very thick, ->] (1.4, 0) -- (2, 0) -- (2, 1.5) -- (1.25, 1.5);
        \node at (2.55, 0.8) {action};
        \node[mybox] at (0, 1.5) {environment};
        \def\x{-3.2}
        \draw[very thick, ->] (-1.2, 1.7) -- (\x, 1.7) -- (\x, 1.1);
        \node at (-2.4, 1.5) {state};
        \node[mybox] at (\x, 0.8) {constraint};
        \draw[very thick, ->] (-2.6, 0.5) -- (-2.6, -0.2) -- (-1.4, -0.2);
        \node at (-2, -0.4) {state};

        \draw[very thick, ->] (-1.2, 1.3) -- (-1.8, 1.3) -- (-1.8, 0.2) -- (-1.4, 0.2);
        \node at (-1.2, 0.8) {reward};
        
        % green check mark
        \draw[green, line width=4pt] (-2.8, 0.2) -- (-2.6, 0) -- (-2.2, 0.4);
        
        \draw[very thick] (-3.7, 0.5) -- (-3.7, 0.2);
        % red cross
        \draw[red, line width=4pt] (-3.9, 0.3) -- (-3.5, -0.1);
        \draw[red, line width=4pt] (-3.9, -0.1) -- (-3.5, 0.3);
    \end{tikzpicture}
    \caption{Illustration of closed-loop constrained RL.}
    \label{fig: closed-loop constrained RL}
\end{figure}

While reinforcement learning (RL)~\citep{RL} is widely successful~\citep{dota2, DQN, AlphaZero}, its application to safety-critical tasks is challenging due to its lack of safety guarantees~\citep{RL_challenges}.
A common approach towards capturing safety in RL is ensuring the satisfaction of safety constraints which prevent a robot from entering unsafe regions \citep{review_safety_levels}. However, verifying that a learned closed-loop policy \emph{never} leads to any constraint violation is in general a nontrivial problem.
To remedy this shortcoming, safe RL has mostly relied on reward shaping to penalize the policy for constraint violations~\citep{review_safety_levels, review_safe_RL, RL_soft_constraint}. At a high level this approach corresponds to imposing \emph{soft} safety constraints and does not provide any guarantees of constraint satisfaction by the closed-loop system~\citep{review_safe_RL}. However, for many safety-critical tasks, like human-robot interaction, autonomous driving~\citep{safe_driving}, or Airborn Collision Avoidance Systems~\citep{ACAS_X}, such safety guarantees are paramount and require maintaining inviolable \textit{hard constraints} in closed loop with the learned policy, as illustrated with Figure~\ref{fig: closed-loop constrained RL}.

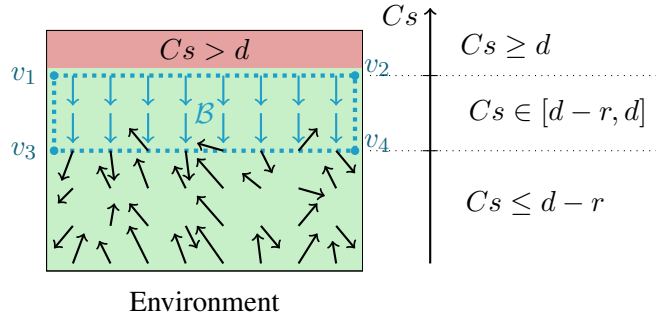
\begin{figure}[t!]
    \centering
    \definecolor{bufferfill}{RGB}{30, 158, 201}
    \definecolor{downboxfill}{RGB}{161, 230, 163}
    \definecolor{myredfill}{RGB}{230, 161, 161}
    \definecolor{bufferfill_accent}{RGB}{19, 101, 128}
    \definecolor{myredfill_accent}{RGB}{193, 51, 51}
    \begin{tikzpicture}[scale=1]
        % state space
        \node at (2, 0.5) {Environment};
        \def\vy{3.5}; % y of the buffer vertices

        %%%%%%%% Rounded corners %%%%%%%%%%%%
        % \draw[rounded corners, very thin, fill=downboxfill!60] (-0.1, 0.9) rectangle (4.1, 4.1);
        % \def\eps{0.07}
        % \filldraw[myredfill, rounded corners] (-\eps, \vy+0.1) -- (-\eps, 4+\eps) -- (4+\eps, 4+\eps) -- (4+\eps, \vy+0.1); % obstacle
        %%%%%%%%%% Square corners %%%%%%%%%%%
        \draw[very thin, fill=downboxfill!60] (-0.1, 0.9) rectangle (4.1, 4.1);
        \filldraw[myredfill] (-0.1, \vy+0.1) -- (-0.1, 4.1) -- (4.1, 4.1) -- (4.1, \vy+0.1); % obstacle
        \draw[very thin] (-0.1, 0.9) rectangle (4.1, 4.1);
        %%%%%%%%%%%%%%%%%%%%%%%%%%%%%%%%%%%%%
        
        \node at (2, 3.85) {$Cs > d$};
        % radius r of the buffer
        \def\r{1.};
        \draw[dotted, bufferfill, ultra thick] (0, \vy) -- (4, \vy) -- (4, \vy-\r) -- (0, \vy-\r) -- (0, \vy); % buffer
        
        \filldraw[bufferfill] (0, \vy) circle (0.05); % v1
        \node at (-0.4, \vy) {\textcolor{bufferfill_accent}{$v_1$}};
        \filldraw[bufferfill] (4, \vy) circle (0.05); % v2
        \node at (4.3, \vy+0.1) {\textcolor{bufferfill_accent}{$v_2$}};
        \filldraw[bufferfill] (4, \vy-\r) circle (0.05); % v3
        \node at (4.3, \vy-\r+0.1) {\textcolor{bufferfill_accent}{$v_4$}};
        \filldraw[bufferfill] (0, \vy-\r) circle (0.05); % v4
        \node at (-0.4, \vy-\r) {\textcolor{bufferfill_accent}{$v_3$}};

        % flow in buffer
        \foreach \x in {0.25, 0.75, ..., 3.75}
            \foreach \y in {\vy-0., \vy-0.5}
                \draw[bufferfill, ->, thick] (\x, \y) -- (\x, \y -0.4);
    
        \node at (2, \vy-\r/2) {\textcolor{bufferfill}{$\mathcal{B}$}};

        % flow below buffer
        \draw[->, thick] (0.25, 2) -- (0.05, 1.8);
        \draw[->, thick] (0.75, 2) -- (0.6, 2.3);
        \draw[->, thick] (1.25, 2) -- (1.1, 2.4);
        \draw[->, thick] (1.75, 2) -- (1.6, 2.3);
        \draw[->, thick] (2.25, 2) -- (1.9, 2.4);
        \draw[->, thick] (2.75, 2) -- (2.5, 2.2);
        \draw[->, thick] (3.25, 2) -- (3.6, 1.9);
        \draw[->, thick] (3.75, 2) -- (3.6, 2.3);
        
        \draw[->, thick] (0.25, 1.5) -- (0., 1.2);
        \draw[->, thick] (0.75, 1.5) -- (0.8, 1.8);
        \draw[->, thick] (1.25, 1.5) -- (1., 1.8);
        \draw[->, thick] (1.75, 1.5) -- (1.9, 1.2);
        \draw[->, thick] (2.25, 1.5) -- (1.9, 1.9);
        \draw[->, thick] (2.75, 1.5) -- (3, 1.2);
        \draw[->, thick] (3.25, 1.5) -- (3.5, 1.8);
        \draw[->, thick] (3.75, 1.5) -- (4., 1.2);

        \draw[->, thick] (0.25, 1) -- (0.4, 1.4);
        \draw[->, thick] (0.75, 1) -- (0.6, 1.3);
        \draw[->, thick] (1.25, 1) -- (1., 1.5);
        \draw[->, thick] (1.75, 1) -- (1.6, 1.4);
        \draw[->, thick] (2.25, 1) -- (1.9, 1.5);
        \draw[->, thick] (2.75, 1) -- (2.7, 1.4);
        \draw[->, thick] (3.25, 1) -- (3.5, 1.4);
        \draw[->, thick] (3.75, 1) -- (3.7, 1.3);
        
        \draw[->, thick] (0.25, 2.5) -- (0.1, 2.1);
        \draw[->, thick] (0.75, 2.5) -- (0.8, 2.1);
        \draw[->, thick] (1.25, 2.5) -- (1., 2.8);
        \draw[->, thick] (1.75, 2.5) -- (1.8, 2.1);
        \draw[->, thick] (2.25, 2.5) -- (1.9, 2.6);
        \draw[->, thick] (2.75, 2.5) -- (2.9, 2.2);
        \draw[->, thick] (3.25, 2.5) -- (3.5, 2.8);
        \draw[->, thick] (3.75, 2.5) -- (4., 2.2);

        \draw[thick, ->] (5, 1) -- (5, 4.4);
        \node at (4.6, 4.3) {$Cs$};
        \node at (6, 3.9) {$Cs \geq d$};
        \draw[dotted] (4.2, \vy) -- (8, \vy);
        \draw[] (4.9, \vy) -- (5.1, \vy);
        \node at (6.7, \vy-\r/2) {$Cs \in [d-r, d]$};
        \draw[dotted] (4.2, \vy-\r) -- (8, \vy-\r);
        \draw[] (4.9, \vy-\r) -- (5.1, \vy-\r);
        \node at (6.4, 1.8) {$Cs \leq d-r$};

        % \node at (2, 4.3) {$\mathcal{S}$};
        % \node at (-0.4, 1.5) {\textcolor{downboxfill!100}{$\mathcal{S}_s$}};
        
    \end{tikzpicture}
    \caption{Schematic illustration of POLICEd RL. To prevent state $s$ from violating an affine constraint represented by $Cs \leq d$, our POLICEd policy enforces $C\dot s \leq 0$ in buffer region $\mathcal{B}$ (\textcolor{bufferfill}{blue}) directly below the unsafe area (\textcolor{myredfill_accent}{red}). The POLICEd policy (arrows in the environment) is affine inside buffer region $\mathcal{B}$ (delimited by vertices $v_1, \hdots, v_4$), which allows us to easily verify whether trajectories can violate the constraint.}
    \label{fig: POLICEd illustration}
\end{figure}

In this paper, we seek to learn a robot control policy guaranteed to satisfy an affine state constraint in a deterministic but black-box environment. The state space region where this constraint is not satisfied corresponds to an unsafe area that must be avoided by the robot. Our key insight is to transform the state space surrounding this unsafe area into a repulsive buffer region as shown in Figure~\ref{fig: POLICEd illustration}. This buffer is made repulsive by learning a policy whose actions push the robot state away from the unsafe area. To enable analytical verification of the repulsive character of our learned policy, we constrain its outputs to be affine over the buffer region. This key property allows to easily guarantee whether an affine constraint is satisfied.

Our proposed framework is agnostic to the choice of the RL training algorithm but relies on its convergence to guarantee safety. Our method is applicable to both systems with continuous and discrete state and action spaces. Additionally, our approach can accommodate black-box environments by using a local measure of their nonlinearity, which can be numerically estimated. To the best of our knowledge, no other work learns a policy in such a way that the closed-loop system is guaranteed to satisfy a hard constraint after training.

To ensure our policy produces affine outputs over the whole buffer region, we draw from existing literature in constraining neural network outputs. While this topic has been widely investigated~\citep{optimization_layer, quadratic_projection, DC3, homogeneous_constraint, Russian_hard_constraint, RAYEN}, we build on the POLICE algorithm proposed in~\citep{police} to constrain our learned policy. Indeed, POLICE is capable of ensuring that a deep neural network with continuous piecewise affine activation functions produces affine outputs over a user-specified region. We build on this work and develop a paradigm for enforcing hard constraints in RL which we call POLICEd RL.

Following this approach, we establish analytical conditions under which our learned policy guarantees constraint satisfaction after training. A natural follow-up question arising from these conditions is whether they are so stringent that no constraint-satisfying policy exist. To answer this crucial problem, we transform the question of existence of such a constraint-satisfying policy into a tractable linear problem. We will then demonstrate through a number of numerical examples how our method can be implemented. Then, using a 7DOF robotic arm, we compare the performance of our method with a number of representative baselines and demonstrate that our method outperforms all the baselines both in terms of its constraint satisfaction as well as its expected accumulated reward.

In this work, we focus on constraining only \emph{actuated states} in an effort to lighten the complexity of our theory. Such a setting is commonly used in a wide range of safety research such as the literature using of control barrier functions~\citep{CBF, ConBaT, hard_soft_barrier, yang2023model} as well as safe RL~\citep{safe_exploration, knuth2021planning, ma2022joint, imagination}. We will provide a detailed review of these works in Section~\ref{subsec: relative degree}.

In summary, our contributions in this work are as follows.
\begin{enumerate}
    \item We introduce POLICEd RL, a novel RL framework that can guarantee satisfaction of hard constraints by the closed-loop system composed of a black-box robot model and a trained policy.
    \item We transform the question of existence of such a constraint-satisfying policy into a tractable linear problem.
    \item We demonstrate the efficiency of our proposed POLICEd RL at guaranteeing the safety of an inverted pendulum and a robotic manipulator in a number of simulations using high-fidelity MuJoCo simulators~\citep{mujoco}.
\end{enumerate}

The remainder of this work is organized as follows. 
In Section~\ref{sec: literature} we provide a survey of related works.
In Section~\ref{sec: prior work} we describe prior work~\citep{police} upon which we build our approach.
In Section~\ref{sec: framework} we introduce our problem formulation along with our framework.
In Section~\ref{sec: constraint} we establish our approach and provide its theoretical guarantees to enforce the satisfaction of affine constraints.
In Section~\ref{sec: implementation} we demonstrate how to implement our proposed POLICEd RL algorithm.
In Section~\ref{sec: simulations} we present several numerical simulations illustrating our approach.
Finally, we conclude the paper in Section~\ref{sec: conclusion}.

\textit{Notation:}
The characteristic function of a set $\mathcal{S} \subseteq \mathbb{R}^n$ is denoted by $\mathbbm{1}_{\mathcal{S}}$. 
The positive integer interval from $a \in \mathbb{N}$ to $b \in \mathbb{N}$ inclusive is denoted by $[\![a, b]\!]$.
Uniform sampling of a variable $x$ in a set $\mathcal{X}$ is denoted by $x \sim \mathcal{U}(\mathcal{X})$.

\section{Related works}\label{sec: literature}

\subsection{Enforcing hard constraints on neural network outputs}

First, we review the literature on enforcing hard constraints on the outputs of deep neural networks (DNNs). Adding a post-processing layer or an extra activation function like $tanh$ or $sigmoid$ to a DNN can easily bound its output. Similarly, linear constraints can be enforced by projecting DNN outputs onto a constraint set using quadratic programming optimization layers \citep{quadratic_projection}. To enforce general convex constraints, \citep{optimization_layer} developed differentiable convex optimization layers that can be incorporated into DNNs.
However, evaluating these optimization layers is computationally expensive, which led to imposing linear constraint without any projection in~\citep{homogeneous_constraint} where the constraint is built into the DNN's architecture. This work was recently extended to enforce affine constraints in localized regions \citep{police}. The main advantage of these two works~\citep{police, homogeneous_constraint} is the absence of computational overhead at deployment where such constrained DNNs become simple multilayer perceptrons.
To go beyond affine constraints, \citep{DC3} used gradient descent along equality constraints until inequality constraints are also satisfied.
Observing that gradient descent can require numerous iterations and suffers convergence issues prompted a more efficient method for hard convex constraint enforcement through the offline computation of feasible sets in \citep{RAYEN}.
In a concurrent work, \citep{Russian_hard_constraint} built DNNs whose outputs satisfy linear and quadratic constraints without solving any optimization problem in the forward pass. All these works enforce constraints on DNN outputs only and do not consider the satisfaction of hard constraints by a trained policy.

\subsection{Constraints in reinforcement learning}

The most common approach to enforce constraints in RL adopts the framework of constrained Markov decision processes (CMDPs)~\citep{CMDP}. CMDPs encourage policies to respect constraints by penalizing the expectation of the cumulative constraint violations \citep{RL_soft_constraint}. Numerous variations of this framework have been developed such as state-wise constrained MDP~\citep{state_wise_constrained_MDP}, constrained policy optimization~\citep{constrained_policy_optimization}, and state-wise constrained policy optimization~\citep{imagination}.
These approaches belong to the category of \textit{soft constraints} as the policy is only \emph{encouraged} to respect the constraint and provides no satisfaction guarantees~\citep{review_safe_RL}.
This category also encompasses work \citep{ConBaT} where a control barrier transformer is trained to avoid unsafe actions, but no safety guarantees can be derived.

\begin{figure}[t!]
    \centering
    \includegraphics[scale=0.8]{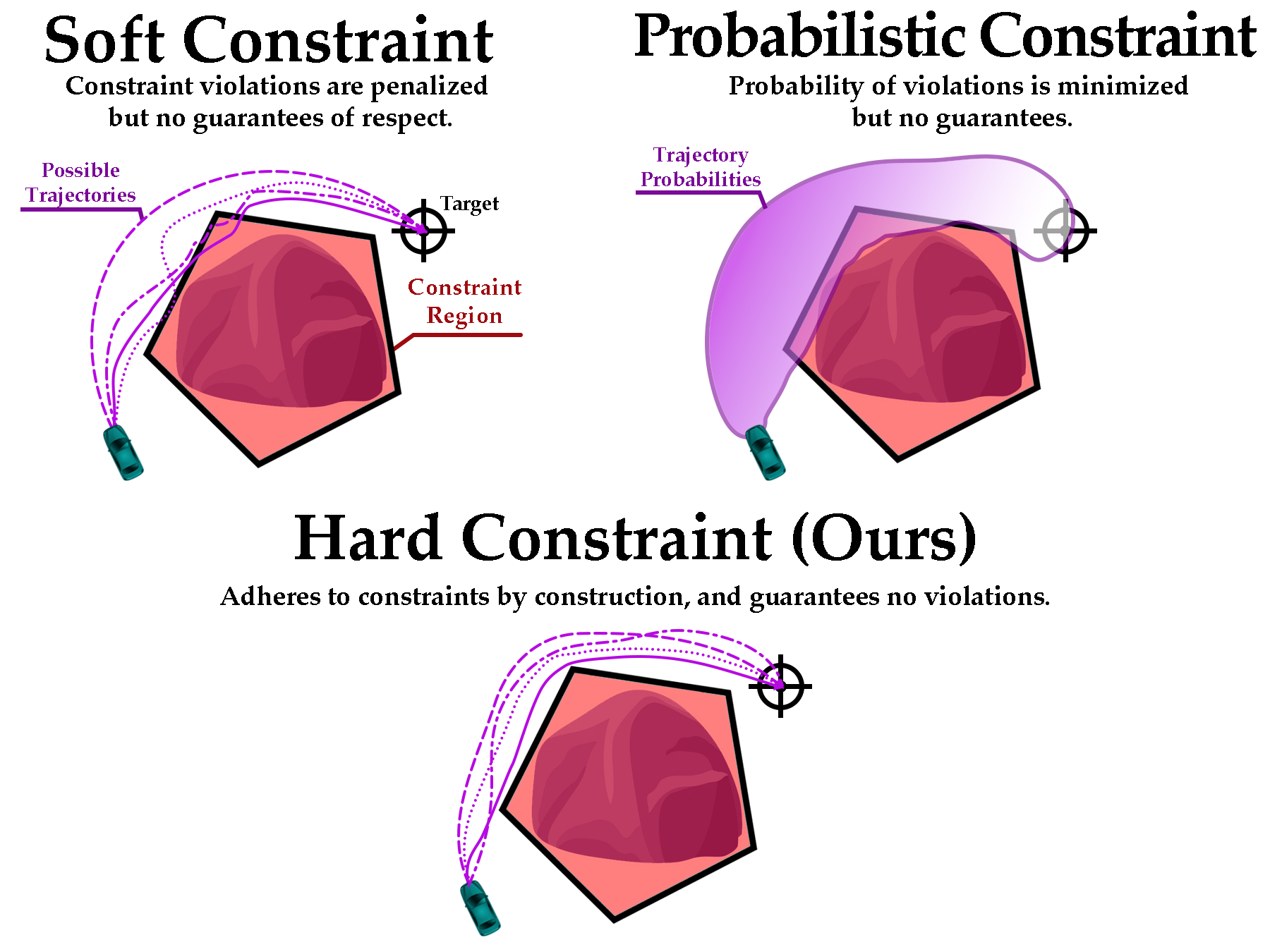}
    \caption{The three categories of constraint satisfaction with increasing guarantees of satisfaction.}
    \label{fig: safety levels}
    \vspace{-1em}
\end{figure}

A \textit{probabilistic} constraint comes with the guarantee of satisfaction with some probability threshold \citep{review_safety_levels} and hence ensures a higher level of safety than soft constraints as shown in Figure~\ref{fig: safety levels}. 
For instance, \citep{probably_approx_correct} derived policies having a high probability of not violating the constraints by more than a small tolerance.
Using control barrier functions, \citep{end_to_end} guaranteed safe learning with high probability.
Since unknown stochastic environments prevent hard constraints enforcement, \citep{hard_soft_barrier} proposed to learn generative model-based soft barrier functions to encode chance constraints.
Similarly, by using a safety index on an unknown environment modeled by Gaussian processes, \citep{probabilistic, knuth2021planning} established probabilistic safety guarantees.

The focus of our work is at the third safety level described in \citep{review_safety_levels} and illustrated in Figure~\ref{fig: safety levels}. This level corresponds to inviolable \emph{hard constraints}.
Works \citep{safe_exploration, Optlayer} both learned safe policies thanks to a differentiable safety layer projecting any unsafe action onto the closest safe action. However, since these safety layers only correct actions, a precise model of the robot dynamics model must be known, which is typically unavailable in RL settings.
This limitation is also shared by~\citep{backward_reachable, BarrierNet}, which require some knowledge of the robot dynamics model to compute either backward reachable sets \citep{backward_reachable} or control barrier functions \citep{BarrierNet}. To circumvent any knowledge of the robot dynamics model, \citep{ISSA} used an implicit black-box model but is restricted to collision avoidance problems in 2D. To avoid these limitations, the Lagrangian-based approach of \citep{ma2022joint} and the model-free approach of \citep{yang2023model} require to a priori specify the functional form of the safety certificate and as they approach zero violations, the cost signal for their DNN approximator becomes sparse, for these reasons they cannot guarantee the validity of their safety certificate.

To sum up, our work differs from all these works as we enforce inviolable \emph{hard} constraints on robot trajectories in closed-loop with a learned control policy while exclusively using a black-box model of the robot dynamics.

\subsection{Relative degree of constraints}\label{subsec: relative degree}

The relative degree of constraints is an important notion in safe RL deserving some introduction.
The \emph{relative degree of a constraint} describes how many times a constraint needs to be differentiated before a control input appears in its expression. For instance, a position constraint has relative degree $1$ for a velocity-controlled robot, but relative degree $2$ for an acceleration-controlled robot. The higher the relative degree, the more inertia the constraint has and the more anticipation is required to respect it. 

Control barrier functions (CBFs), one of the most common tools to enforce constraints, require constraints of relative degree 1~\citep{CBF}. Albeit not being mentioned, the relative degree of the constraints studied in \citep{ConBaT} has to be 1 for its CBF theory to apply \citep{CBF}. Indeed, both of the experiments of \citep{ConBaT} consist in avoiding obstacles with cars actuated directly by their steering angle, so their relative degree is 1.
Similarly, work \citep{hard_soft_barrier} relies on CBF and hence assumes implicitly constraints of relative degree 1.
Only recently have CBFs been specifically extended to handle constraints of higher relative degrees with, for instance, exponential CBFs~\citep{Exponential_CBF}, generalized CBFs~\citep{ma2021model}, and high order CBFs (HOCBF)~\citep{HOCBF}. However, these CBF methods require knowledge of the robot dynamics model, which is typically not available in RL. 

Work \citep{knuth2021planning} proposes to model the system dynamics as fully-actuated and affine so that all constraints are of relative degree 1.
Work \citep{safe_exploration} mentions constraints of higher relative degree through the notion of inertia, but only address them experimentally by giving a larger margin to correct trajectories before they violate the constraint.
The paper \citep{ma2022joint} only mentions high order constraints to avoid dealing with them by excluding the states leading to violation, i.e., when their relative degree is higher than 1 from the safe set. 
Similarly, \citep{ISSA} uses a Lagrangian approach with state-dependent multipliers and a constraint of relative degree 2. However, the method of \citep{ISSA} is designed for 2D systems with sufficient control power to effectively counter their inertia, hence circumventing the major challenge of high relative degree constraints.

We follow the common approach of most of the safe RL literature~\citep{safe_exploration, knuth2021planning, ma2022joint, ConBaT, hard_soft_barrier, imagination} and choose to study constraints of relative degree 1 in this work.

\subsection{Black-box safety with control theory}

While CBFs are prevalent in safe RL~\citep{ConBaT, Sablas, hard_soft_barrier, yang2023model}, other approaches backed by control theory have also investigated safety in black-box RL. 
Most notably, model predictive control (MPC) is particularly well-suited to bring safety guarantees to RL settings~\cite{hewing2020learning}. MPC schemes predict safe and optimal control inputs in a receding-horizon fashion by using a model of the dynamics to anticipate their behavior~\cite{hewing2020learning}. Implementing MPC for black-box systems typically requires learning either a robust~\cite{aswani2013provably, di2004multi} or stochastic~\cite{lorenzen2017stochastic} model of the dynamics. Then, the theory of robust~\cite{robustMPC} and stochastic control~\cite{wabersich2021probabilistic} can provide safety guarantees despite the uncertainty or disturbances in the learned models~\cite{hewing2020learning}.
Similarly to CBFs, MPC can also serve as safety filter to enforce constraints on learned policies~\citep{hewing2020learning, wabersich2021predictive}. While robust control tends to be overly conservative in practice~\cite{weakRobustControl} and stochastic MPC offers limited safety guarantees~\cite{weakStochasticMPC}, the main limitation of black-box MPC is the unavoidable and notoriously computationally difficult receding-horizon optimization problem to be solved at each time step~\cite{hewing2020learning}.

More closely related to this work is the literature on optimal control of constrained piecewise affine systems~\cite{christophersen2007piecewise}. Indeed, nonlinear dynamics can be locally approximated as piecewise affine to develop robust controllers with provable performance guarantees~\cite{rebuttal_Martin, rebuttal_Schuppen} while enforcing linear temporal logic specifications~\cite{rebuttal_Belta}. However, these works assume knowledge of the robot dynamics.

\section{Prior Work}\label{sec: prior work}

\begin{figure}[htb!]
    \centering
    \includegraphics[scale=0.7]{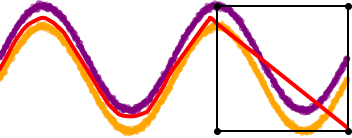}
    \caption{Classification task of \textcolor{orange}{orange} versus \textcolor{purple}{purple} by a learned decision boundary (\textcolor{red}{red}) which is required to be affine inside the black square. POLICE~\citep{police} guarantees the DNN is affine in the region of interest.}
    \label{fig: affine classification}
\end{figure}

In this section, we review prior work \citep{police} upon which we build our approach.
The POLICE algorithm \citep{police} utilizes the spline theory of DNNs~\citep{balestriero18b} to modify the training process of a DNN by guaranteeing that the DNN outputs are strictly affine within a user-specified region as shown in Figure~\ref{fig: affine classification}.

More specifically, consider a DNN represented by a function $f_\theta : \mathbb{R}^{d_0} \rightarrow \mathbb{R}^{d_L}$ composed of $L$ layers of width $d_1, ..., d_L$ defined as $f_\theta := f_{\theta_L}^{(L)} \circ \hdots \circ f_{\theta_1}^{(1)}$. Let layer $i \in [\![1,L]\!]$ take the form $f_{\theta_i}^{(i)}(x) = \sigma \big( W^{(i)}x + b^{(i)} \big)$, where $\sigma$ is a pointwise activation function, $W^{(i)} \in \mathbb{R}^{d_{i} \times d_{i-1}}$ is a weight matrix, and $b^{(i)} \in \mathbb{R}^{d_i}$ is a bias vector. The layer parameters are gathered into $\theta_i := \{ W^{(i)},\ b^{(i)}\}$ while the later parameters themselves are gathered as $\theta := \{\theta_1, \hdots, \theta_L\}$.

The framework of \citep{police} assumes that nonlinearities $\sigma$ are continuous piecewise affine functions such as (leaky-)ReLU, absolute value or max-pooling. This assumption is common in the field of DNN verification \citep{neural_Lyap, weng2018towards, zhang2018efficient, zhang2019recurjac}.

Under this assumption, $f_\theta$ is a composition of affine functions $W^{(i)}x + b^{(i)}$ and piecewise affine functions $\sigma$. As a result, the DNN $f_\theta$ itself is also piecewise affine. Hence, we can partition input space $\mathbb{R}^{d_0}$ into $M$ regions $\mathcal{R}_j$, $j \in [\![1, M]\!]$, over which $f_\theta$ is affine. The nonlinearities of $f_\theta$ occur at the boundaries between regions $\mathcal{R}_j$. Then,
\begin{equation}\label{eq: CPA}
    f_\theta(x) = \sum_{j = 1}^M \big( D_j x + e_j \big) \mathbbm{1}_{x\, \in\, \mathcal{R}_j},
\end{equation}
where $D_j \in \mathbb{R}^{d_L \times d_0}$, and $e_j \in \mathbb{R}^{d_L}$ are per-region slope and offset matrices, and regions $\mathcal{R}_j \subseteq \mathbb{R}^{d_0}$ are such that $\bigcup_{j=1}^M \mathcal{R}_j = \mathbb{R}^{d_0}$. These regions $\mathcal{R}_j$ are defined by their characteristic functions $\mathbbm{1}_{x\, \in\, \mathcal{R}_j}$.
The formulation of \eqref{eq: CPA} is discussed at length in \citep{activation_pattern}, which also describes how to compute regions $\mathcal{R}_j$ and matrices $D_j$ and $e_j$ based on the parameter vector $\theta$.

The POLICE algorithm \citep{police} builds on this formulation to impose one user-specified polytopic region $\mathcal{R}_{\text{user}}$ into the partition of input space $\mathbb{R}^{d_0}$. As a result, DNN $f_\theta$ becomes affine on this specific region $\mathcal{R}_{\text{user}}$.
This affine region is enforced by an extra bias term $b_{\text{extra}}$ calculated for each layer such that the whole region $\mathcal{R}_{\text{user}}$ possesses the same activation pattern. Note that standard unconstrained optimization is still used to train the DNN.
Once training is over, a POLICEd DNN behaves similarly to a standard DNN as the affine region enforcement is guaranteed by the stored bias of each layer. Thus, POLICEd DNNs have no overhead at test time compared to standard unconstrained DNNs \citep{police}.

\section{Framework}\label{sec: framework}

Let us now introduce the framework of our problem of interest. We consider a robot of state $s$ with deterministic dynamics modeled by a continuous function $f$ and of form
\begin{equation}\label{eq: nonlinear dynamics}
    \dot s(t) = f\big( s(t), a(t) \big), \qquad a(t) \in \mathcal{A}, \qquad s(0) \sim \rho_0,
\end{equation}
where $a$ is the action input and $\rho_0$ is the distribution of initial states.
The assumption of deterministic dynamics is common in the literature of safe RL with hard constraints as exhibited by \citep{safe_exploration, ma2022joint, backward_reachable, BarrierNet, yang2023model, ISSA, imagination}.
Let state space $\mathcal{S}$ be a compact convex polytope of $\mathbb{R}^n$ and admissible action set $\mathcal{A}$ be a compact convex subset of $\mathbb{R}^m$. Similar assumptions are adopted by \citep{ma2022joint, backward_reachable, BarrierNet}.
The state of the robot is constrained by a single affine inequality 
\begin{equation}\label{eq: constraint}
    y(t) := C s(t) \leq d, \qquad \text{for all}\ t \geq 0,
\end{equation}
with $C \in \mathbb{R}^{1 \times n}$ and $d \in \mathbb{R}$.
The robot starts from an initial state $s(0) \sim \rho_0$. At every time instant $t \geq 0$, the robot observes its state $s(t) \in \mathcal{S}$ and takes an action according to its deterministic feedback policy $a(t) = \mu_\theta\big( s(t) \big) \in \mathcal{A}$ modeled by a DNN $\mu_\theta$ parameterized by $\theta$. Then, the robot receives a reward $R\big( s(t), a(t)\big)$.
Our objective is to train policy $\mu_\theta$ to maximize the expected discounted reward over trajectories generated by policy $\mu_\theta$ while respecting constraint~\eqref{eq: constraint}:
\begin{equation}\label{eq: expected reward}
    \underset{\theta}{\max}\, \mathcal{G}(\mu_\theta) :=  \underset{s_0 \sim \rho_0}{\mathbb{E}} \int_0^{\infty}  \gamma^t R\big( s(t), \mu_\theta(s(t))\big) dt \quad \text{s.t.}\ \eqref{eq: constraint},
\end{equation}
where $\gamma \in (0,1]$ is a discount factor. We emphasize that constraint~\eqref{eq: constraint} is a \textit{hard constraint} to be respected at all times.
We can now formally define our problem of interest.

\begin{problem}\label{prob: continuous}
    Given:
    \begin{enumerate}[left=0pt]
        \item a black box continuous control system in the form of \eqref{eq: nonlinear dynamics};
        \item a compact convex polytopic state space $\mathcal{S} \subset \mathbb{R}^n$;
        \item a compact convex admissible action set $\mathcal{A} \subset \mathbb{R}^m$;
        \item an affine hard constraint in the form of \eqref{eq: constraint};
        \item a DNN policy $\mu_\theta\big( s(t) \big)$ parameterized by $\theta$;
    \end{enumerate}
    our goal is to solve
    \begin{align*}%\label{eq: optimization problem}
        \theta^* = \underset{\theta}{\arg\max}\ \mathcal{G}(\mu_\theta) \quad \text{s.t.} \quad \dot s(t) = f\big( s(t), \mu_\theta(s(t)) \big), \quad s(0) \sim \rho_0, \quad Cs(t) \leq d, \quad \text{for all}\ t \geq 0.
    \end{align*}
\end{problem}

Our approach focuses on deterministic dynamics with no model mismatch, although it can be readily extended to uncertain dynamics thanks to robust safe control \citep{robust_safe_control}. Additionally, we consider the robot dynamics model $f$ to be an implicit black-box, meaning that we can evaluate $f$ but we do not have access to the equations or analytical form of $f$. This is similar to the online RL setting where $f$ is a simulator or where $f$ is the actual robot.

\section{Constrained Reinforcement Learning}\label{sec: constraint}

In this section, we establish a method to solve Problem~\ref{prob: continuous}. To enforce the satisfaction of affine constraint $Cs \leq d$, we construct a repulsive buffer where $C\dot s \leq 0$ around the constraint line $Cs = d$ as illustrated in Figure~\ref{fig: POLICEd illustration}. This repulsive buffer will then guarantee that closed-loop trajectories of the robot cannot breach the constraint.
We will establish the analytical safety guarantees of our method in this section and then in the next section, we discuss the implementation details of our approach.

\subsection{Guaranteed satisfaction of hard constraints}

We start by constructing a repulsive buffer in front of the constraint violation line defined by~\eqref{eq: constraint}. Let $r > 0$ be the `radius' of this buffer defined as
\begin{equation}\label{eq: buffer}
    \mathcal{B} := \big\{ s \in \mathcal{S} : C s \in [d - r, d]\big\}.
\end{equation}
Note that any state trajectory initially verifying constraint~\eqref{eq: constraint}, i.e., $Cs(0) \leq d$, has to cross buffer $\mathcal{B}$ before being able to violate the constraint. Therefore, if buffer $\mathcal{B}$ cannot be crossed, then constraint~\eqref{eq: constraint} cannot be violated. To design a policy $\mu_\theta$ incapable of crossing buffer $\mathcal{B}$, we need the following result. 

\begin{lemma}\label{lemma: buffer polytope}
    Buffer $\mathcal{B}$ of \eqref{eq: buffer} is a polytope.
\end{lemma}
\begin{proof}
    By definition~\eqref{eq: buffer}, buffer $\mathcal{B}$ can be written as $\mathcal{B} = C^{-1}\big([d - r, d]\big) \cap \mathcal{S}$, where $C^{-1}\big([d - r, d]\big) := \big\{s : Cs \in [d - r, d]\big\}$ denotes the inverse image of the interval $[d - r, d]$. The inverse image of a set is always defined, even if matrix $C$ is not invertible.
    Therefore, $\mathcal{B}$ is the intersection of affine variety $C^{-1}\big([d - r, d]\big)$ and polytope $\mathcal{S}$ and hence $\mathcal{B}$ is a polytope according to result 4 of section 3.1 of \citep{Convex_Polytopes}.
\end{proof}
Then, buffer $\mathcal{B}$ has a finite number $N$ of vertices gathered in the set $\mathcal{V}\big(\mathcal{B}\big) := \big\{v_1, \hdots, v_{N} \big\}$.

We choose a deterministic policy modeled by a POLICEd DNN $\mu_\theta : \mathcal{S} \rightarrow \mathcal{A}$. We adopt the framework of \citep{police} as discussed in Section~\ref{sec: prior work}. We assume that the activation functions of $\mu_\theta$ are continuous piecewise affine functions such as (leaky) ReLU, absolute value, or max-pooling.
As mentioned in Section~\ref{sec: prior work}, this POLICEd DNN architecture allows the user to specify a polytopic region $\mathcal{R}_{\text{user}}$ of the state space $\mathcal{S}$ where the outputs of $\mu_\theta$ are affine. We choose $\mathcal{R}_{\text{user}} = \mathcal{B}$ as illustrated in Figure~\ref{fig: POLICEd illustration}.

Following \eqref{eq: CPA}, the affine character of $\mu_\theta$ over region $\mathcal{B}$ is equivalent to the existence of matrices $D_{\theta} \in \mathbb{R}^{m \times n}$ and $e_{\theta} \in \mathbb{R}^m$ such that
\begin{equation}\label{eq: police}
    \mu_\theta(s) = D_{\theta} s + e_{\theta} \quad \text{for all}\ s \in \mathcal{B}.
\end{equation}

Having an affine policy on $\mathcal{B}$, we would like to couple it with affine robot dynamics to obtain a simple constraint enforcement process. However, in general, robot dynamics~\eqref{eq: nonlinear dynamics} are nonlinear. We will thus use an affine approximation of the robot dynamics inside buffer $\mathcal{B}$ using the following definition.

\begin{definition}\label{def: epsilon}
    An \emph{approximation measure} $\varepsilon$ of dynamics~\eqref{eq: nonlinear dynamics} with respect to constraint~\eqref{eq: constraint} and buffer~\eqref{eq: buffer} is any $\varepsilon \geq 0$ for which there exists any matrices $A \in \mathbb{R}^{n \times n}$, $B \in \mathbb{R}^{n \times m}$, and $c \in \mathbb{R}^n$ such that
    \begin{equation}\label{eq: approximation}
        \big| C f(s, a) - C(A s + B a + c) \big| \leq \varepsilon,
    \end{equation}
    for all $s \in \mathcal{B}$, and all $a \in \mathcal{A}$.
\end{definition}

Despite this approximation, note that we will guarantee the satisfaction of constraint~\eqref{eq: constraint} with actual dynamics~\eqref{eq: nonlinear dynamics}. 
To help us easily compute a value for $\varepsilon$ using system identification techniques \citep{system_id} like linear least square \citep{linear_least_squares}, we take advantage of the following property of approximation measures.

\begin{lemma}\label{lemma: epsilon}
    Given dynamics $f$ of~\eqref{eq: nonlinear dynamics}, constraint matrix $C$ of~\eqref{eq: constraint} and buffer $\mathcal{B}$ of~\eqref{eq: buffer}, any $\varepsilon$ sufficiently large is an approximation measure in the sense of Definition~\ref{def: epsilon}. 
\end{lemma}
\begin{proof}
    We want to prove that any $\varepsilon$ sufficiently large is an approximation measure in the sense of Definition~\ref{def: epsilon}. We will first determine the infimum of all approximation measures before proving that any $\varepsilon$ larger than this infimum is an approximation measure.

    For every matrices $A \in \mathbb{R}^{n \times n}$, $B \in \mathbb{R}^{n \times m}$, and $c \in \mathbb{R}^n$ introduce
    \begin{equation*}\label{eq: epsilon A B c}
        \varepsilon^*_{ABc} := \min \left\{ \varepsilon : \big|Cf(s,a) -C(As + Ba + c)\big| \leq \varepsilon, \quad \text{for all}\ a \in \mathcal{A},\ \text{and all}\ s \in \mathcal{B} \right\}.
    \end{equation*}
    Note that the minimum in $\varepsilon^*_{ABc}$ exists since the function to minimize is continuous ($f$ is assumed continuous at \eqref{eq: nonlinear dynamics}) and sets $\mathcal{A}$ and $\mathcal{B}$ are compact.
    Then, all $\varepsilon^*_{ABc}$ are approximation measures as they satisfy Definition~\ref{def: epsilon}. Define
    \begin{equation*}
        \varepsilon^* := \inf \big\{ \varepsilon^*_{ABc} : A \in \mathbb{R}^{n \times n},\ B \in \mathbb{R}^{n \times m},\ c \in \mathbb{R}^n \big\}.
    \end{equation*}
    Let us now prove that the infimum of all approximation measures is $\varepsilon^*$, by first showing that $\varepsilon^*$ is their largest lower bound.
	If $\varepsilon$ is an approximation measure, then it has a triplet $(A,B,c)$ satisfying \eqref{eq: approximation}. By definition of $\varepsilon^*_{ABc}$ and $\varepsilon^*$, we then have $\varepsilon \geq \varepsilon^*_{ABc} \geq \varepsilon^*$.
	Thus, $\varepsilon^*$ is a lower bound to all approximation measures.
	
	Let $\alpha$ be a lower bound to all approximation measures.
	Since all the $\varepsilon^*_{ABc}$ are approximation measures, $\alpha \leq \varepsilon^*_{ABc}$ for all $A, B, c$, i.e., $\alpha \leq \varepsilon^*$ by definition of $\varepsilon^*$.
	Therefore, $\varepsilon^*$ is the largest lower bound of all approximation measures.
    
    We will now show that any $\varepsilon$ larger than $\varepsilon^*$ is in fact an approximation measure.
    Let $\varepsilon > \varepsilon^*$. Since $\varepsilon^*$ is the infimum of all approximation measures, there exists an approximation measure $\alpha$ such that $\varepsilon^* \leq \alpha < \varepsilon$.
	Since $\alpha$ is an approximation measure, there exists a triplet $(A,B,c)$ for which \eqref{eq: approximation} holds with upper bound $\alpha$.
    Note that \eqref{eq: approximation} also holds for this same triplet but with upper bound $\varepsilon$ since $\varepsilon > \alpha$.
    Then $\varepsilon$ is an approximation measure.
\end{proof}

Note that intuitively, the value of $\varepsilon$ quantifies the quality of the approximation over buffer $\mathcal{B}$ of possibly nonlinear robot dynamics~\eqref{eq: nonlinear dynamics} by affine system
\begin{equation}\label{eq: affine dynamics}
    \dot s(t) = A s(t) + B a(t) + c, \quad a(t) \in \mathcal{A}.
\end{equation}
We will show how to \emph{guarantee} satisfaction of constraint \eqref{eq: constraint} by black-box environment~\eqref{eq: nonlinear dynamics} armed \emph{only} with an approximation measure $\varepsilon$ and \emph{without} knowing matrices $A$, $B$, $c$ or dynamics $f$.

We define the safe states as $\mathcal{S}_s := \big\{ s \in \mathcal{S} : C s < d \big\}$. We only consider trajectories remaining in state space $\mathcal{S}$, which we define as $\tau^{\mathcal{S}}\big(s_0, a(\cdot) \big) := \big\{ s(t) : s(t) \in \mathcal{S}\ \text{and follows}\ \eqref{eq: nonlinear dynamics} \big\}$ for all $s_0 \in \mathcal{S}$ and adequate actions $a(\cdot) \in \mathcal{A}$.
We can now state our central result relating repulsive buffer $\mathcal{B}$ to trajectories $\tau^{\mathcal{S}}$ continuously respecting the constraints.

\begin{theorem}\label{thm: admissible trajectories}
    If for some approximation measure $\varepsilon$, repulsion condition 
    \begin{equation}\label{eq: repulsion}
        C f\big(v, \mu_\theta(v)\big) \leq -2\varepsilon,
    \end{equation}
    holds for all $v \in \mathcal{V}\big(\mathcal{B}\big)$, then $\tau^{\mathcal{S}}(s_0, \mu_\theta) \subseteq \mathcal{S}_s$ for all $s_0 \in \mathcal{S}_s$.
\end{theorem}
\begin{proof}
    The intuition behind this proof is to use \eqref{eq: repulsion} and approximation~\eqref{eq: approximation} to show that $C\dot s \leq 0$ for all $s \in \mathcal{B}$, which in turn prevents trajectory from crossing buffer $\mathcal{B}$ and hence from violating the constraint.

    Since $\varepsilon$ is an approximation measure, there exists matrices $A$, $B$ and $c$ verifying \eqref{eq: approximation}. We can now extend repulsion condition~\eqref{eq: repulsion} from the vertices of buffer $\mathcal{B}$ to the whole set $\mathcal{B}$. For $v \in \mathcal{V}\big(\mathcal{B}\big)$,
    \begin{align}\label{eq: vertex repulsion}
        C \big( Av + B\mu_\theta(v) + c \big) &\leq \big|C\big( Av + B\mu_\theta(v) + c - f(v, \mu_\theta(v))\big)\big| + C f\big(v, \mu_\theta(v)\big) \nonumber \\
        &\leq \varepsilon + C\dot v \leq \varepsilon -2\varepsilon \leq -\varepsilon,
    \end{align}
    where the first inequality is a triangular inequality, the second follows from \eqref{eq: approximation} and \eqref{eq: nonlinear dynamics}, and the third ineqaulity follows from \eqref{eq: repulsion}.
    Using the fact that an affine function is uniquely determined by its values on the vertices of a full-dimensional polytope \cite{Convex_Polytopes}, \cite{rebuttal_Schuppen}, we will show that \eqref{eq: vertex repulsion} is also valid all over polytope $\mathcal{B}$ and not just at its vertices $\mathcal{V}\big(\mathcal{B}\big)$. More specifically, since $\mathcal{B}$ is the convex hull of its vertices $\{v_1, \hdots, v_{N}\} = \mathcal{V}\big(\mathcal{B}\big)$ \cite{Convex_Polytopes}, for all $s \in \mathcal{B}$, there exists $\alpha_1, \hdots, \alpha_{N} \in \mathbb{R}$ such that $\alpha_k \geq 0$, $\sum_{k = 1}^{N} \alpha_k = 1$ and $s = \sum_{k=1}^{N} \alpha_k v_k$. Then, \eqref{eq: police} yields
    \begin{align}\label{eq: vertices eps}
        C \big( As + B\mu_\theta(s) + c \big) &= C \big( As + B(D_{\theta} s + e_{\theta}) + c \big) = C(A + B D_{\theta})s + C(B e_{\theta} + c) \nonumber  \\
        &= C (A + B D_{\theta})\sum_{k=1}^{N} \alpha_k v_k + C (B e_{\theta} + c)\sum_{k = 1}^{N} \alpha_k \nonumber  \\
        &= \sum_{k = 1}^{N} \alpha_k C \big( (A + B D_{\theta})v_k + B e_{\theta} + c \big)  \\
        &= \sum_{k = 1}^{N} \alpha_k C \big( A v_k + B \mu_\theta(v_k) + c \big) \leq \sum_{k = 1}^{N} \alpha_k (-\varepsilon) = -\varepsilon, \nonumber
    \end{align}
    where the inequality comes from \eqref{eq: vertex repulsion} applied on each vertex $v_k$. For any state $s \in \mathcal{B}$, we have
    \begin{align}\label{eq: nonpositive derivative}
        C \dot s &= C f \big(s, \mu_\theta(s) \big) \leq \big|C \big( f(s, \mu_\theta(s)) - A s - B\mu_\theta(s) - c \big) \big| + C\big( A s + B \mu_\theta(s) + c \big) \nonumber \\
        &\leq \varepsilon - \varepsilon \leq 0,
    \end{align}
    where we first use the triangular inequality, then \eqref{eq: approximation} and \eqref{eq: vertices eps}.
    Having proved \eqref{eq: nonpositive derivative}, we will now show that it prevents all trajectories $\tau^{\mathcal{S}}(s_0, \mu_\theta)$ from exiting safe set $\mathcal{S}_s$ when $s_0 \in \mathcal{S}_s$.
    
    We prove this by contradiction. Assume that trajectory $\tau^{\mathcal{S}}(s_0, \mu_\theta) \nsubseteq \mathcal{S}_s$ for some $s_0 \in \mathcal{S}_s$. Since $\tau^{\mathcal{S}}(s_0, \mu_\theta) \subset \mathcal{S}$, there exists some $T > 0$ such that $s(T) \in \mathcal{S} \backslash \mathcal{S}_s$. Then, by definition of $\mathcal{S}_s$, $y(T) = C s(T) \geq d$. Since $s_0 \in \mathcal{S}_s$, $y(0) = C s_0 < d$. By continuity of the function $y$, there exists a time $t_2 \in (0, T]$ at which $y(t_2) = C s(t_2) = d$. Let $t_0 \geq 0$ be the last time at which $\tau^{\mathcal{S}}(s_0, \mu_\theta)$ entered $\mathcal{B}$, so that $s(t) \in \mathcal{B}$ for all $t \in [t_0, t_2]$.

    Note that $y(t) = Cs(t)$ is continuously differentiable since dynamics~\eqref{eq: nonlinear dynamics} are continuous.
    Then, according to the Mean Value Theorem \citep{Analysis}, there exists $t_1 \in (t_0, t_2)$ such that $\dot y(t_1) = \big( y(t_2) - y(t_0) \big)/(t_2 - t_0)$. %\frac{y(t_2) - y(t_0)}{t_2 - t_0}$.
    By construction of $t_0$ and $t_2$, we have $t_2 - t_0 > 0$ and $y(t_2) - y(t_0) > 0$. Therefore, $\dot y(t_1) = C \dot s(t_1) > 0$ with $s(t_1) \in \mathcal{B}$, which contradicts \eqref{eq: nonpositive derivative}. Therefore, all trajectories $\tau^{\mathcal{S}}(\cdot, \mu_\theta)$ starting in safe set $\mathcal{S}_s$ remain in $\mathcal{S}_s$.
\end{proof}

Theorem~\ref{thm: admissible trajectories} guarantees that trajectories remaining in $\mathcal{S}$ and steered by $a(t) = \mu_\theta\big( s(t) \big)$ satisfy constraint~\eqref{eq: constraint} at all times as long as repulsion condition~\eqref{eq: repulsion} is satisfied. This condition \eqref{eq: repulsion} guarantees that trajectories cannot cross buffer $\mathcal{B}$ and hence cannot violate constraint~\eqref{eq: constraint}.

\begin{remark}\label{rmk: epsilon}
    In order to guarantee satisfaction of constraint~\eqref{eq: constraint} through Theorem~\ref{thm: admissible trajectories}, we need an approximation measure $\varepsilon$. In practice, we can use data-driven techniques for estimating $\varepsilon$. Our safety guarantees are valid as long as our estimator $\tilde{\varepsilon}$ overshoots the true $\varepsilon$ since that will help make $\tilde{\varepsilon}$ an approximation measure according to Lemma~\ref{lemma: epsilon}. \footnote{However, in practice an excessively large value of $\varepsilon$ will be detrimental to the learning performance of $\mu_\theta$ since repulsion condition~\eqref{eq: repulsion} will become harder to enforce as $\varepsilon$ increases.}
\end{remark}

Theorem~\ref{thm: admissible trajectories} highlights the major strength of POLICEd RL: to ensure constraint satisfaction we only need to check whether \eqref{eq: repulsion} holds at the vertices of $\mathcal{B}$. 
% If $\mathcal{B}$ is a hyper-rectangle in $\mathbb{R}^n$, we need to check $N = 2^n$ points. 
Without POLICE, $\mu_\theta$ would not be affine over $\mathcal{B}$, and repulsion condition~\eqref{eq: repulsion} would need to be verified at numerous intermediary points depending on the smoothness of $\mu_\theta$ and the size of the buffer. Such an approach is similar to the $\delta$-completeness guarantees of \citep{neural_Lyap, delta_complete} and would be much more computationally intensive than our approach.

Theorem~\ref{thm: admissible trajectories} and more specifically Condition \eqref{eq: repulsion} implicitly assume that the robot has access to $\dot v$, the derivative of the state, which is usually not the case in typical RL environments \citep{Gym}. To address this, we provide a straightforward extension of Theorem~\ref{thm: admissible trajectories} to the widespread framework of discrete-time setting.
We define the discrete-time state of the robot as
\begin{equation}\label{eq: discrete observations}
    s_{j+1} := s_j + f(s_j, a_j)\delta t, 
\end{equation}
for all $s_j \in \mathcal{S}$, $a_j \in \mathcal{A}$, $j \in \mathbb{N}$, and some time step $\delta t > 0$. The discrete trajectory of policy $\mu_\theta$ starting from $s_0 \in \mathcal{S}$ and remaining in state space $\mathcal{S}$ is defined as
\begin{equation}\label{eq: discrete traj}
    \tau_d^{\mathcal{S}}(s_0, \mu_\theta) := \big(s_0, s_1, \hdots\big) \in \mathcal{S}^{\mathbb{N}},
\end{equation}
such that $s_{j+1} = s_j + f(s_j, \mu_\theta(s_j))\delta t \in \mathcal{S}$ for all $j \in \mathbb{N}$.

\begin{cor}\label{cor: discrete repulsion}
    If for some approximation measure $\varepsilon$, the following discrete repulsion condition holds 
    \begin{equation}\label{eq: discrete repulsion}
        C \big( s_{j+1} - s_j \big) \leq -2\varepsilon \delta t \quad \text{for all} \ s_j \in \mathcal{V}\big(\mathcal{B}\big),
    \end{equation}
    with $a_j = \mu_\theta(s_j)$, and buffer $\mathcal{B}$ is wide enough such that
    \begin{equation}\label{eq: buffer width}
        \max \big\{ C \big( s_{j+1} - s_j \big) : s_j \in \mathcal{S}_s,\ a_j = \mu_\theta(s_j)  \big\} \leq r,
    \end{equation}
    then $\tau_d^{\mathcal{S}}(s_0, \mu_\theta) \subset \mathcal{S}_s$ for all $s_0 \in \mathcal{S}_s$.
\end{cor}
\begin{proof}
    As in the proof of Theorem~\ref{thm: admissible trajectories}, we take advantage of the linearity of dynamics~\eqref{eq: affine dynamics} to extend \eqref{eq: discrete repulsion} from the vertices of $\mathcal{B}$ to the whole set $\mathcal{B}$. Note that for any $s_t \in \mathcal{V}\big(\mathcal{B}\big)$, combining repulsion condition \eqref{eq: discrete repulsion} with \eqref{eq: approximation} yields
    \begin{align*}
        C \big( A s_t + B\mu_\theta(s_t) + c \big) &\leq \big|C\big( A s_t + B\mu_\theta(s_t) + c \big) - C f\big(s_t, \mu_\theta(s_t)\big)\big| + C f\big(s_t, \mu_\theta(s_t)\big) \\
        &\leq \varepsilon +\frac{1}{\delta t} C \big( s_{t+1} - s_t\big) \leq \varepsilon -2\varepsilon \delta t \frac{1}{\delta t} \leq -\varepsilon.
    \end{align*}
    As in the proof of Theorem~\ref{thm: admissible trajectories}, we use the convexity of polytope $\mathcal{B}$ of vertices $\mathcal{V}\big(\mathcal{B}\big)$ and the linearity of approximation~\eqref{eq: affine dynamics} to obtain 
    \begin{equation}\label{eq: cor 1 proof 1}
        C \big( A s + B \mu_\theta(s) + c \big) \leq -\varepsilon, \quad \text{for all} \ s \in \mathcal{B}. 
    \end{equation}
    We can now revert this inequality to the discrete dynamics. For $s_t \in \mathcal{B}$,
    \begin{align}\label{eq: nonpositive discrete derivative}
        C \frac{s_{t+1} - s_t}{\delta t}  &\leq \left|C \frac{s_{t+1} - s_t}{\delta t} - C \big( A s_t + B \mu_\theta(s_t) + c \big) \right| + C \big( A s_t + B \mu_\theta(s_t) + c \big) \nonumber \\
        &\leq \big|C f\big(s_t, \mu_\theta(s_t)\big) - C \big( A s_t + B \mu_\theta(s_t) + c \big) \big| -\varepsilon \leq \varepsilon - \varepsilon \leq 0.
    \end{align}
    Note that the first inequality follows from the triangular inequality, the second from the definition of $s_{t+1}$ and \eqref{eq: cor 1 proof 1}, and the third inequality stems from \eqref{eq: approximation}.
    We will now show that \eqref{eq: nonpositive discrete derivative} prevents all trajectories $\tau_d^{\mathcal{S}}(s_0, \mu_\theta)$ from exiting safe set $\mathcal{S}_s$ when $s_0 \in \mathcal{S}_s$.
    
    We assume for contradiction purposes that trajectory $\tau_d^{\mathcal{S}}(s_0, \mu_\theta) \notin \mathcal{S}_s$ for some $s_0 \in \mathcal{S}_s$. Since $\tau_d^{\mathcal{S}}(s_0, \mu_\theta) \in \mathcal{S}$, there exists some $t \in \mathbb{N}$ such that $s_t \in \mathcal{S}_s$ and $s_{t+1} \in \mathcal{S} \backslash \mathcal{S}_s$. Thus, $C s_t < d$ and $C s_{t+1} \geq d$. 
    Now \eqref{eq: buffer width} yields
    \begin{equation*}
        C (s_{t+1} - s_t) \leq r,
    \end{equation*}
    since $s_t \in \mathcal{S}_s$ and $\mu_\theta(s_t) \in \mathcal{A}$. Then, $C s_t \geq C s_{t+1} -r \geq d -r$. Thus, $s_t \in \mathcal{B}$. Since $C s_t < d$ and $C s_{t+1} \geq d$, we have
    \begin{equation*}
        C \frac{s_{t+1} - s_t}{\delta t} > 0,
    \end{equation*}
    which contradicts \eqref{eq: nonpositive discrete derivative}. Therefore, all trajectories $\tau_d^{\mathcal{S}}$ starting in safe set $\mathcal{S}_s$ remain in $\mathcal{S}_s$.
\end{proof}

Note that condition~\eqref{eq: buffer width} requires the buffer to be wide enough not to be `jumped' over in a single time step by the discrete dynamics. Condition~\eqref{eq: discrete repulsion} makes buffer $\mathcal{B}$ repulsive so that discrete trajectories $\tau_d^{\mathcal{S}}$ cannot leave safe set $\mathcal{S}_s$.

\subsection{Existence conditions}

While Theorem~\ref{thm: admissible trajectories} and Corollary~\ref{cor: discrete repulsion} provide a way to verify the satisfaction of hard constraint~\eqref{eq: constraint}, they do not entirely solve Problem~\ref{prob: continuous}. A natural remaining piece is the following existence condition.

\begin{problem}\label{prob: existence}
    Under what conditions does there exist an admissible policy $\mu_\theta$ satisfying Theorem~\ref{thm: admissible trajectories}?
\end{problem}
Indeed, if the conditions of Theorem~\ref{thm: admissible trajectories} cannot be satisfied, then training $\mu_\theta$ will fail to solve Problem~\ref{prob: continuous}, which is why Problem~\ref{prob: existence} is crucial. To address this issue, we first reformulate the conditions of Theorem~\ref{thm: admissible trajectories} into a tractable existence problem.

\begin{proposition}\label{prop: equivalence condition}
    Finding an admissible policy $\mu_\theta$ satisfying Theorem~\ref{thm: admissible trajectories} is equivalent to finding a matrix $D_{\theta} \in \mathbb{R}^{m \times n}$ and a vector $e_{\theta} \in \mathbb{R}^m$ which satisfy the following conditions for all vertices $v \in \mathcal{V}\big(\mathcal{B}\big)$
    \begin{subequations}
        \begin{align}
        &C \big( (A + B D_{\theta})v + B e_{\theta} + c \big) \leq -\varepsilon, \label{eq: linear repulsion}\\
        &D_{\theta} v + e_{\theta} \in \mathcal{A}. \label{eq: admissible} 
        \end{align}
    \end{subequations}
\end{proposition}
\begin{proof}    
    Following definition \eqref{eq: police} of policy $\mu_\theta$, \eqref{eq: linear repulsion} is equivalent to \eqref{eq: vertex repulsion}.
    Similarly, a policy $\mu_\theta$ satisfying Theorem~\ref{thm: admissible trajectories} also verifies \eqref{eq: vertex repulsion}.
    
    The only statement left to prove is the equivalence between \eqref{eq: admissible} and policy $\mu_\theta$ being admissible over buffer $\mathcal{B}$.
    Admissibility is formalized as $\mu_\theta(s) \in \mathcal{A}$ for all $s \in \mathcal{B}$, which combined with \eqref{eq: police}, trivially implies \eqref{eq: admissible}. 
    
    Conversely, assume \eqref{eq: admissible} holds and let us prove that $\mu_\theta$ is admissible. Let $s \in \mathcal{B}$. Since $\mathcal{V}\big(\mathcal{B}\big) = \{v_1, \hdots, v_{N}\}$ are the vertices of convex set $\mathcal{B}$, there exists $\alpha_1, \hdots, \alpha_{N} \in \mathbb{R}$ such that $\alpha_k \geq 0$, $\sum_{k = 1}^{N} \alpha_k = 1$ and $s = \sum_{k=1}^{N} \alpha_k v_k$. Then, \eqref{eq: police} yields
    \begin{align*}
        \mu_\theta(s) = D_{\theta} s + e_{\theta} = D_{\theta} \left( \sum_{k=1}^{N} \alpha_k v_k \right) + e_{\theta} \left( \sum_{k=1}^{N} \alpha_k \right) = \sum_{k=1}^{N} \alpha_k (D_{\theta} v_k + e_{\theta}).
    \end{align*}
    Then, \eqref{eq: admissible} coupled with the convexity of $\mathcal{A}$ yields $\mu_\theta(s) \in \mathcal{A}$ for all $s \in \mathcal{B}$, hence completing the proof.   
\end{proof}

Proposition~\ref{prop: equivalence condition} translates the feasibility of Theorem~\ref{thm: admissible trajectories} into two sets of $N$ linear conditions to verify. Additionally, if the input set $\mathcal{A}$ is a polytope, \eqref{eq: admissible} can be simplified to a linear existence problem. Hence, the question of the existence of a policy $\mu_\theta$ satisfying Theorem~\ref{thm: admissible trajectories} can be answered efficiently with linear programming.
Notice that \eqref{eq: linear repulsion} admits plenty of solutions as long as $C B \neq 0$. This observation is in fact related to the concept of \emph{relative degree} whose definition we now formalize.

\begin{definition}\label{def: relative degree}
    The \emph{relative degree} $\gamma$ of system~\eqref{eq: nonlinear dynamics} with output~\eqref{eq: constraint} is the order of its input-output relationship, i.e., $\gamma := \min\big\{ r \in \mathbb{N} : \frac{\partial}{\partial a} \frac{\partial^r y(t)}{\partial t^r} \neq 0 \big\}$.
\end{definition}
In simpler words, the relative degree is the minimal number of times output $y$ has to be differentiated until input $a$ appears.

\begin{proposition}\label{prop: relative degree}
    If affine dynamics~\eqref{eq: affine dynamics} with output~\eqref{eq: constraint} have a relative degree $1$, then condition~\eqref{eq: linear repulsion} admits solutions. Otherwise, \eqref{eq: linear repulsion} admits solutions if and only if $c_v^* \leq -\varepsilon$ where 
    $$c_v^* := \max\big\{ CA v + Cc : v \in \mathcal{V}\big(\mathcal{B}\big) \big\}.$$
\end{proposition}
\begin{proof}
    Following Definition~\ref{def: relative degree} with dynamics~\eqref{eq: affine dynamics} and constraint~\eqref{eq: constraint}, we calculate
    \begin{align*}
        \frac{\partial}{\partial a} \frac{\partial y(t)}{\partial t} = \frac{\partial}{\partial a} \frac{\partial Cs(t)}{\partial t} = \frac{\partial}{\partial a} C \dot s(t) = \frac{\partial}{\partial a} C \big(As(t) + Ba(t) + c\big) = CB.
    \end{align*}
    Then, a relative degree $1$ entails $CB \neq 0$, i.e., there exists $j \in [\![1, m]\!]$ such that the $j^{th}$ component of $CB$ is non-zero. Then, we choose $e_{\theta}$ to be the zero vector of $\mathbb{R}^m$ except for its $j^{th}$ component to be $\frac{-c_v^* -\varepsilon}{[CB]_j}$, so that $CB e_{\theta} = -c_v^* -\varepsilon$. We also choose $D_{\theta} = 0 \in \mathbb{R}^{m \times n}$.
    Then, for all $v \in \mathcal{V}\big(\mathcal{B}\big)$ the left-hand side of \eqref{eq: linear repulsion} simplifies to
    \begin{equation*}
        C A v  - c_v^* -\varepsilon + C c \leq c_v^* - c_v^* -\varepsilon = -\varepsilon,
    \end{equation*}
    where the first inequality comes from the definition of $c_v^*$. Therefore, $(D_{\theta}, e_{\theta})$ is a solution to \eqref{eq: linear repulsion}.

    On the other hand, if dynamics~\eqref{eq: affine dynamics} with output~\eqref{eq: constraint} have relative degree larger than $1$, then $CB = 0$, i.e., the policy has no direct impact on output $y$. In that case, \eqref{eq: linear repulsion} simplifies to $C(Av +c) \leq -\varepsilon$, i.e., $c_v^* \leq -\varepsilon$.
\end{proof}

To check the existence of a policy satisfying Theorem~\ref{thm: admissible trajectories}, we cannot rely solely on the hope that the constraint will enforce itself. Thus, in practice, we need $C B \neq 0$, i.e., the relative degree of system~\eqref{eq: affine dynamics} must be $1$.

Since affine system~\eqref{eq: affine dynamics} is only an approximation of nominal dynamics~\eqref{eq: nonlinear dynamics}, they do not necessarily have the same relative degree. However, such a preventable discrepancy only hampers practical implementation of Theorem~\ref{thm: admissible trajectories} by rendering $\varepsilon$ larger than necessary.
Indeed, work \citep{feedback_linearization} discusses how to infer the relative degree of a black box system from first principles. Hence, affine approximation~\eqref{eq: affine dynamics} can and should be designed to match the relative degree of nominal dynamics~\eqref{eq: nonlinear dynamics} to make $\varepsilon$ as small as possible. This reasoning prompts the following practical consideration.

\begin{remark}\label{rmk: relative degree}
    In practice, to find a policy satisfying Theorem~\ref{thm: admissible trajectories}, output~\eqref{eq: constraint} of system~\eqref{eq: nonlinear dynamics} needs relative degree $1$.
\end{remark}

The intuition behind Remark~\ref{rmk: relative degree} is that the derivative of the constrained states must be actuated to allow immediate corrective actions. A relative degree $1$ is also required by CBFs \citep{CBF} and by numerous works in constrained RL \citep{knuth2021planning, ma2022joint, ConBaT, hard_soft_barrier} as we already discussed in Section~\ref{subsec: relative degree}. Let us now detail the implementation of our method.

\section{Implementation}\label{sec: implementation}
\begin{figure}[htb!]
    \centering
    \includegraphics[scale = 0.7]{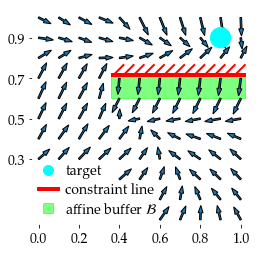}
    \caption{State space $\mathcal{S}$ with arrows denoting state transitions under POLICEd policy $\mu_\theta$ for linear environment~\eqref{eq: toy dynamics}. The affine buffer $\mathcal{B}$ (\textcolor{green!50!black}{green}) pushes states away from the constraint line (\textcolor{red}{red}) before heading towards the target (\textcolor{cyan}{cyan}).}
    \label{fig: toy policed}
\end{figure}

In this section, we demonstrate how our method can be implemented in a continuous 2D environment where a point mass moves according to the following discrete-time dynamics
\begin{equation}\label{eq: toy dynamics}
    s(t+\delta t) = s(t) + a(t) \delta t,
\end{equation}
with actions $a(t) \in \mathcal{A} := [-1, 1]^2$, states $s(t) \in \mathcal{S} := [0, 1]^2$ and time step $\delta t = 0.1$. We assume dynamics~\eqref{eq: toy dynamics} to be a black-box from the controller's perspective. The reward signal is proportional to the distance between the state and a target state located at $(0.9, 0.9)$, as illustrated in Figure~\ref{fig: toy policed}.

We consider the line joining states $(0.4, 0.7)$ to $(1., 0.7)$ to form an obstacle. We will guarantee that trajectories do not cross this constraint line from below by placing our buffer right under it, as shown in Figure~\ref{fig: toy policed}. Note that a policy trained to reach the target will not cross the constraint line from above as that would move the state away from the target as seen in Figure~\ref{fig: toy policed}. Thus, we only need to prevent crossings from below. The constraint is parameterized by $C = [0,1]$ and $d=0.7$ in the notations of~\eqref{eq: constraint}, i.e., $s_2 \leq 0.7$ when $s_1 \geq 0.4$.

The first step of our approach is to estimate the size of our buffer. To do so, we uniformly sample states $s(t) \sim \mathcal{U}(\mathcal{S})$, actions $a(t) \sim \mathcal{U}(\mathcal{A})$, and the corresponding next states $s(t+\delta t)$ from~\eqref{eq: toy dynamics} to be stored in a dataset $\mathcal{D}$. With this dataset we can calculate the minimal buffer radius $r$ following condition~\eqref{eq: buffer width} and we find that for this setting $r = 0.1$, which we can analytically verify with \eqref{eq: toy dynamics} and \eqref{eq: buffer width} as follows:
\begin{equation*}
    C\big( s(t+\delta t) - s(t) \big) = s_2(t+\delta t) - s_2(t) = \delta t \big(-s_2(t) + a_2(t)\big) \leq \delta t \big(-0 + 1) = \delta t
\end{equation*}

Following \eqref{eq: buffer} we define buffer $\mathcal{B}$ delimited by vertices $\mathcal{V}(\mathcal{B}) = \big\{(0.4, 0.7), (1, 0.7), (1, 0.6), (0.4, 0.6)\big\}$ which will be provided to our algorithm to learn a POLICEd policy $\mu_\theta$.

The next step is to determine an approximation measure $\varepsilon$ of the system's nonlinearity. Since dynamics~\eqref{eq: toy dynamics} are linear, affine approximation \eqref{eq: approximation} is exact with $\varepsilon = 0$. 

Finally, we train the POLICEd policy $\mu_\theta$ to ensure repulsion condition~\eqref{eq: discrete repulsion} holds, i.e., $s_2(t+\delta t) - s_2(t) \leq 0$ for $s(t) \in \mathcal{V}(\mathcal{B})$. This is illustrated by the arrows pointing down in the affine buffer of Figure~\ref{fig: toy policed}.
As suggested in \citep{police}, once training is completed, we create a copy $\kappa_\theta$ of the POLICEd DNN $\mu_\theta$, where the POLICEd extra-bias $b_{\text{extra}}$ (discussed at the end of Section~\ref{sec: prior work}) are directly embedded into the standard bias of $\kappa_\theta$. Thus, copy $\kappa_\theta$ is a standard DNN without overhead.
We summarize the POLICEd RL process in Algorithm~\ref{alg: POLICEd RL}.

\begin{algorithm}[htbp!]
\caption{POLICEd RL} \label{alg: POLICEd RL}
\begin{algorithmic}[1]
\Require Environment~\eqref{eq: discrete observations}, constraint~\eqref{eq: constraint}, transition dataset $\big(s, a, s') \sim \mathcal{D} \subset \mathcal{S} \times \mathcal{A} \times \mathcal{S}$

\State Calculate buffer radius $r$ with \eqref{eq: buffer width} from dataset $\mathcal{D}$
\State Determine buffer $\mathcal{B}$ and its vertices $\mathcal{V}(\mathcal{B})$ with \eqref{eq: buffer}
\State Sample transitions $(s,a,s') \sim \mathcal{D}$ s.t. $s \in \mathcal{B}$ and use least-square approximation to get $\varepsilon$ from \eqref{eq: approximation}
\State Train the POLICEd RL agent $\mu_\theta$ until repulsion condition~\eqref{eq: discrete repulsion} holds on the polytopic buffer's vertices $\mathcal{V}(\mathcal{B})$
\State Copy $\mu_\theta$ into a standard DNN $\kappa_\theta$ of identical weights and biases increased by the POLICEd extra-bias $b_{\text{extra}}$ of $\mu_\theta$
\Ensure Trajectories $\tau_d^{\mathcal{S}}(s_0, \mu_\theta)$ of \eqref{eq: discrete traj} starting from safe state $s_0 \in \mathcal{S}_s$ do not leave safe set $\mathcal{S}_s$ and copied policy $\kappa_\theta = \mu_\theta$ has no inference-time overhead.
\end{algorithmic}
\end{algorithm}

\section{Simulations}\label{sec: simulations}

We will now test our POLICEd RL framework through more challenging realistic simulations to answer the following questions:
\begin{itemize}%[\itemindent=-10pt]
    \item[\textbf{Q1}] Can POLICEd RL produce safer policies than a baseline?
    \item[\textbf{Q2}] Can POLICEd RL generate policies achieving higher rewards than a baseline, while guaranteeing constraint satisfaction?
    \item[\textbf{Q3}] Can POLICEd RL be expanded to higher-dimensional systems with realistic dynamics?
\end{itemize}

\subsection{Inverted pendulum experiment}\label{subsec: pendulum}
\begin{figure}[htb!]
    \centering
    \includegraphics[scale=0.3]{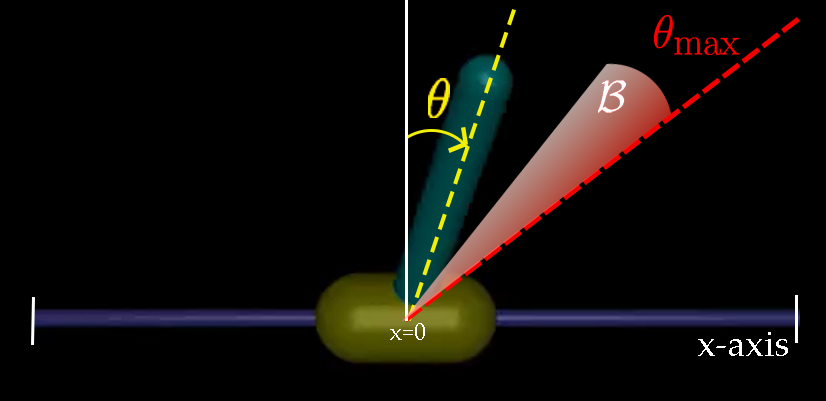}
    \caption{The inverted pendulum Gym environment \citep{Gym} annotated with cart position $x$, pendulum angle $\theta$, and buffer $\mathcal{B}$.}
    \label{fig: inverted pendulum}
    \vspace{-1em}
\end{figure}

We begin by testing POLICEd RL on the OpenAI inverted pendulum environment~\citep{Gym} which uses the MuJoCo physics engine~\citep{mujoco}. The inverted pendulum environment has a 4-dimensional observation space with cart position $x$, pendulum angle $\theta$, linear cart velocity $\dot x$, and angular velocity $\dot \theta$ as illustrated in Figure~\ref{fig: inverted pendulum}. The action space is the force applied to the cart $a$ proportional to $\ddot x$.

We investigate different safety constraints to prevent the pendulum from falling. First, we consider a constraint of relative degree 1 where we want to enforce $\dot \theta(t) \leq 0$ for $\theta(t) \in [0.1, 0.2]\, rad$ to push $\theta$ away from its upper limit $\theta_{max} = 0.2\, rad$. With state $s(t) = \big(x(t), \theta(t), \dot x(t), \dot \theta(t)\big)$, the constraint is $C s(t) \leq d$ with $d = 0$ and $C = \big[0\ 0\ 0\ 1\big]$.
We now need to create a buffer wide enough such that it cannot be `jumped' over by the robot in a single time-step. Following~\eqref{eq: buffer}, we build a buffer of width $r$ as $\mathcal{B} = \big\{ (x, \theta, \dot x, \dot \theta) : x \in [-0.9, 0.9],\ \theta \in [0.1, 0.2],\ \dot x \in [-1, 1],\ \dot \theta \in [-r, 0] \big\}$. Buffer $\mathcal{B}$ stays clear off $x = \pm 1$ since these locations cause a large state discontinuity preventing stabilization. 
Following step 1 of Algorithm~\ref{alg: POLICEd RL}, we uniformly sample states $s \sim \mathcal{U}\big(\mathcal{S}\big)$, actions $a \sim \mathcal{U}\big(\mathcal{A}\big)$, corresponding next state $s'$ and we use \eqref{eq: buffer width} to compute $r \approx 1.03$ for action magnitudes $|a| \leq 1$.
To guarantee $\dot \theta \leq 0$, Theorem~\ref{thm: admissible trajectories} leads us to enforce $\ddot \theta \leq 0$ for all states in buffer $\mathcal{B}$.

We can now compute an approximation measure $\varepsilon$ from~\eqref{eq: approximation}. We uniformly sample states $s \sim \mathcal{U}\big(\mathcal{B}\big)$, actions $a \sim \mathcal{U}\big(\mathcal{A}\big)$, we get the corresponding next state $s'$ and approximate $\dot s \approx (s' - s)/ \delta t$ with $\delta t = 0.02\, s$ which is the environment time-step. A least-square fit following \eqref{eq: approximation} yields a value $\varepsilon \approx 0.7$.

Note that training to satisfy a constraint is often adversarial to accomplishing the task, and thus often causes training instability. Throughout our development process, we discovered techniques to greatly improve training time and sample complexity, which are further discussed in Appendix~\ref{sec: training tips}. For example, with the inverted pendulum scenario, we train policy $\mu_\theta$ by iterating over two phases. The first phase is a standard RL training where the initial state is regularly reset either around the origin or inside the affine buffer. Once a reward threshold is met, the \emph{constraint training} phase starts. The state is iteratively reset to all the vertices of buffer $\mathcal{B}$ and the trajectory is propagated for a single time step to evaluate whether repulsion condition~\eqref{eq: discrete repulsion} holds. Only the experiences where~\eqref{eq: discrete repulsion} is \emph{not} verified are added to the replay memory with negative rewards in order to promote the respect of~\eqref{eq: discrete repulsion}. After several rounds of these updates, the first training phase resumes and the process begins anew until condition~\eqref{eq: discrete repulsion} holds everywhere on the vertices of $\mathcal{B}$ and the maximal reward is achieved. We summarize this training process in Algorithm~\ref{alg: training}.

\begin{figure}[htb!]
    \centering
    \includegraphics[scale=0.6]{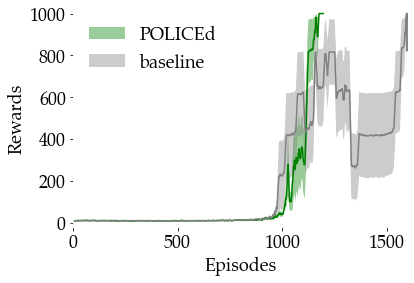}
    \caption{Reward curves for the inverted pendulum (max=1000). The solid lines correspond to the average and the shaded regions to the $95\%$ confidence interval over 5 runs.}
    \label{fig:reward curve}
\end{figure}

We use Proximal Policy Optimization (PPO)~\citep{PPO} to learn both a baseline and a POLICEd policy. The baseline is a standard PPO policy that does not have the enforced affine buffer $\mathcal{B}$ of the POLICEd policy. They both follow the same training procedure described above in the same environment where they receive identical penalties for constraint violations. Each episode has a maximal length of 1000 time steps with a reward of $1$ if the pole is upright and $0$ otherwise. The reward curves of Figure~\ref{fig:reward curve} show that both methods achieve maximal reward.

\begin{figure}[htb!]
    \centering
    \includegraphics[scale=0.6]{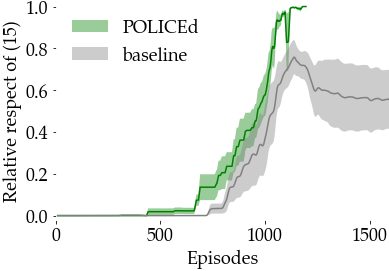}
    \caption{Relative portion of buffer $\mathcal{B}$ where repulsion condition~\eqref{eq: discrete repulsion} is respected. The baseline never succeeds in entirely enforcing~\eqref{eq: discrete repulsion} hence allowing possible constraint violations. The solid lines correspond to the average and the shaded regions to the $95\%$ confidence interval over 5 runs.}
    \label{fig:respect curve}
\end{figure}

During training we measure the proportion of buffer $\mathcal{B}$ where repulsion condition~\eqref{eq: discrete repulsion} is satisfied and report these proportions in Figure~\ref{fig:respect curve}. Since the POLICEd policy achieves maximal reward while completely satisfying repulsion condition~\eqref{eq: discrete repulsion} at episode 1180 it stops training, whereas Figure~\ref{fig:respect curve} shows that the baseline never succeeds in enforcing \eqref{eq: discrete repulsion} over the entire buffer. Moreover, the constraint training phase causes a large drop in the baseline rewards as seen in Figure~\ref{fig:reward curve}. Our POLICEd policy guarantees the satisfaction of constraint $\dot\theta \leq 0$ by enforcing repulsion condition~\eqref{eq: discrete repulsion}, i.e., $\ddot \theta \leq 0$. Consequently, it guarantees that trajectories starting from $\dot \theta(0) < 0$ will never reach any state where $\dot \theta(t) \geq 0$.

We will now empirically evaluate the performance of POLICEd RL at ensuring satisfaction of a constraint of relative degree 2. Although our theory cannot provide any safety guarantees, we will show that POLICEd RL performs better and is safer than the baseline. 

\begin{figure}[htb!]
    \centering
    \subfloat[Baseline]{\includegraphics[scale=0.6]{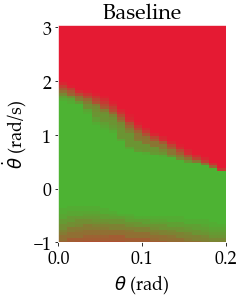}}\hspace{2cm}
    \subfloat[POLICEd]{\includegraphics[scale=0.6]{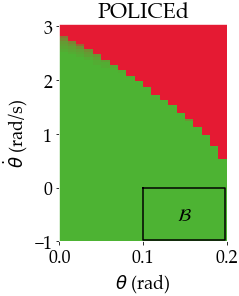}}
    \caption{Success rates for the policies to maintain $|\theta| \leq 0.2\, rad$ for the inverted pendulum given an initial state $\big(0, \theta, 0, \dot\theta \big)$. \textcolor{green!70!black}{Green}: stabilized. \textcolor{red!80!black}{Red}: failure. The black box shows buffer $\mathcal{B}$ where the POLICEd policy guarantees $\ddot \theta \leq 0$. Position constraint $|\theta| \leq 0.2\, rad$ is of relative degree 2. Some initial conditions pictured here \emph{cannot be stabilized} since the controller does not have direct control action on $\theta$, only on $\dot \theta$. However, we see that our POLICEd policy maintains safety in a larger region of the state space than the baseline.}
    \label{fig:failure rates}
\end{figure}

We want the pendulum to maintain $|\theta| \leq \theta_{\max} = 0.2\, rad$ (See Figure~\ref{fig: inverted pendulum}). This position constraint is of relative degree $2$ since the action (pushing the cart) only impacts directly the angular acceleration $\ddot \theta$. Our empirical evaluation consists in resetting the state of the inverted pendulum at a variety of initial conditions $\big(\theta, \dot\theta \big)$, and seeing how often policies fail to maintain $|\theta| \leq 0.2\, rad$. 
Some initial conditions \emph{cannot be stabilized} since the controller does not have direct control action on $\theta$, only on $\dot \theta$. Indeed $\theta(\delta t) = \theta_0 + \delta t\, \dot \theta_0 > 0.2\, rad$, if $\theta_0 = 0.2 \, rad$ and $\dot \theta_0 > 0$ not matter the control action.
We present our results in Figure~\ref{fig:failure rates} where we can see that the POLICEd policy stabilizes a span of initial states much larger than both buffer $\mathcal{B}$ and the ones stabilized by the baseline. 

Then, POLICED RL has shown to be empirically effective in increasing the set of stabilizable initial conditions even with constraints of high relative degree.
Our results indicate that both \textbf{Q1} and \textbf{Q2} are answered positively as POLICEd RL produces safer policies achieving higher rewards than our baseline in the inverted pendulum environment.

\subsection{Robotic arm}\label{subsec: arm}
\begin{figure}[t!]
    \centering
    % \hfill
    \subfloat{\includegraphics[width=0.49\columnwidth]{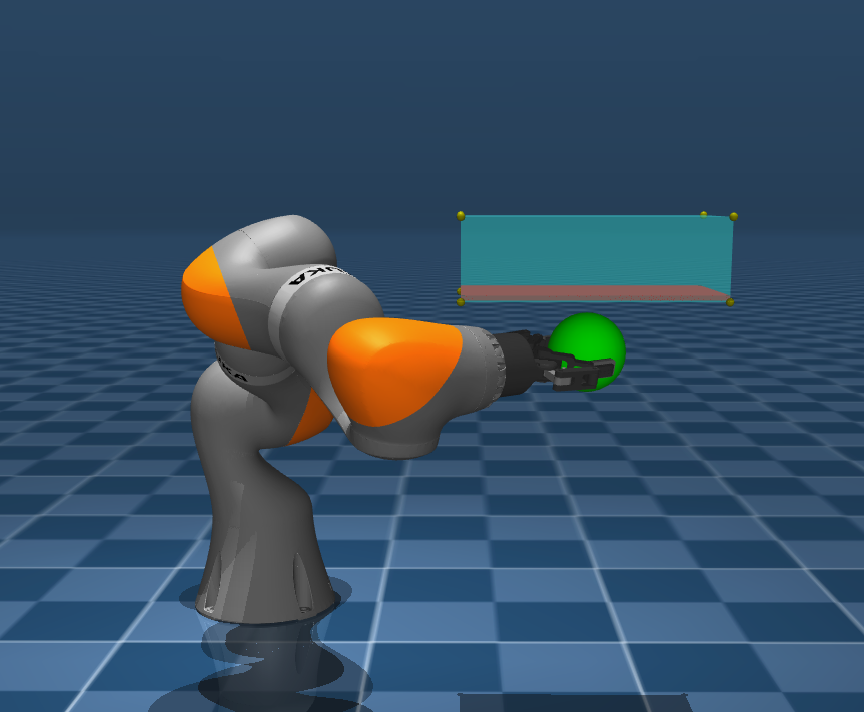}}\hfill
    \subfloat{\includegraphics[width=0.49\columnwidth]{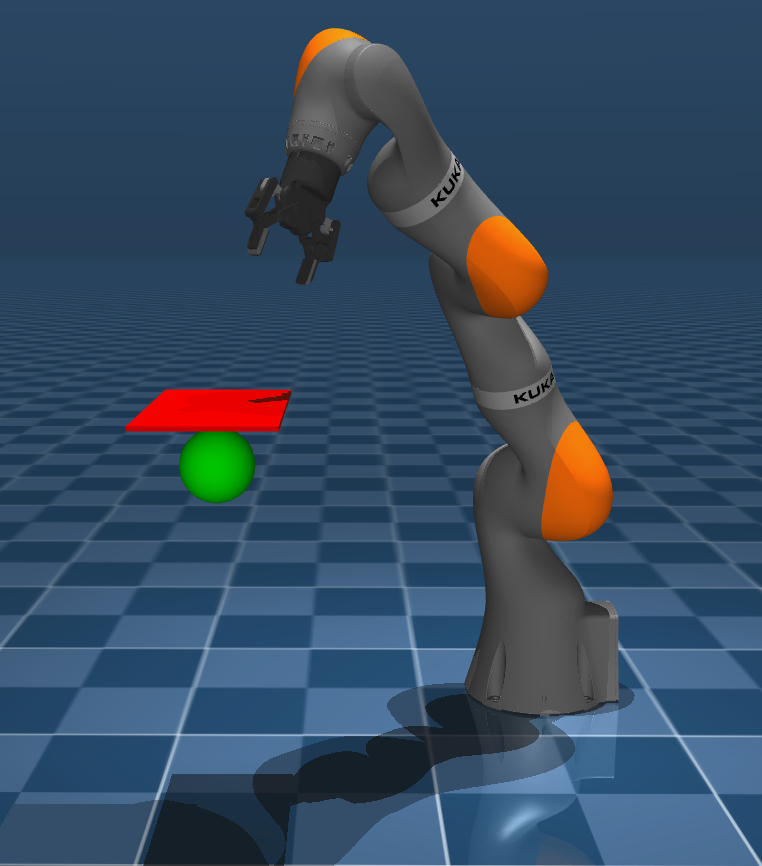}}%\hfill
    \caption{Robotic arms tasked to bring their end-effectors to the target (\textcolor{green!80!black}{green}) while avoiding the constraint surface (\textcolor{red}{red}). The POLICEd method uses a buffer region (\textcolor{cyan}{cyan}). The arms are shown in two random configurations.}
    \label{fig: safe KUKA arms}
    \vspace{-1em}
\end{figure}

We further showcase our method on a robotic arm with $7$ degrees of freedom. To implement our approach, we rely on the high-fidelity MuJoCo simulator \citep{mujoco}. We developed a custom environment where the robotic arm aims to reach a 3D target with its end-effector while avoiding an unsafe region as illustrated in Figure~\ref{fig: safe KUKA arms}.
More specifically, we define the state space of the environment $s \in \mathbb{R}^{10}$ as the arm's joint angles $s_{joints} \in \mathbb{R}^{7}$ and the X-Y-Z coordinates of the end-effector $s_{end} \in \mathbb{R}^3$. At each timestep, the policy provides a change in joint angles $a \in \mathbb{R}^{7}$. The target is kept constant across all episodes. At the start of each episode, the starting state of the arm is uniformly sampled from the joint space.

At each timestep, the robotic arm is assigned a reward function $R(s,a)$ which is composed of four terms: the straight line distance $R_d(s)$ between the end-effector and the target, a bonus $R_b(s,a)$ for reaching the target, a penalty $R_{\text{inf}}(a)$ for generating an infeasible joint state or exiting the state space, and a penalty $R_{\text{unsf}}(s,a)$ for violating the constraint.

We perform an ablation study to showcase the necessity of our POLICEd method to ensure both high rewards and constraint satisfaction. To begin, we learn a POLICEd policy using Twin Delayed DDPG (TD3)~\citep{TD3}, an improved version of Deep Deterministic Policy Gradient (DDPG)~\citep{DDPG}. After training each model, we deployed the fully-trained policy onto 500 episodes on our Safe Arm Task which is illustrated in Figure~\ref{fig: safe KUKA arms}. From these episodes we collected a series of metrics which we summarize in Table~\ref{tab: comparison}:
\begin{enumerate}%[left=0pt]
    \item \emph{Completion \%:} The percentage of episodes where the policy reaches the target to evaluate task completion rate.
    \item \emph{Completion \% w/o Violation:} The percentage of episodes where the policy reaches the target \emph{without} violating the constraint, to evaluate its safety capabilities.
    \item \emph{Average Reward:} The average reward earned to evaluate how efficiently the policy completes the task.
    \item \emph{Average Constraint Satisfaction:} The percentage of episodes where the constraint is never violated to evaluate the safety capabilities of the policy.
\end{enumerate}

We proceeded with the ablation study by training and evaluating variations of our initial policy where we either removed our POLICE module or the reward penalties. We began by training a traditional TD3 policy without our POLICE module and evaluating it as described. We then switched to a reward function without $R_{\text{unsf}}$ and trained a TD3 policy without both the POLICE module and without the constraint violation penalties. We also trained a POLICEd policy without the penalty term in order to showcase how reward shaping is necessary to tune the affine region of the policy to avoid the constraint region.

\begin{table*}[htb!]
    \definecolor{tablegray}{RGB}{90, 98, 115}
    \centering
    \begin{tabular}{ccccc}
        \toprule Models & Completion \% & \begin{tabular}{@{}c@{}}Completion \% \\ w/o Violation \end{tabular} & \begin{tabular}{@{}c@{}} Average reward \\ $\pm 95$\% CI \end{tabular} & \begin{tabular}{@{}c@{}}Average Constraint \\ Satisfaction $\pm 95$\% CI \end{tabular} \\ \midrule
        \begin{tabular}{@{}c@{}}
        \textcolor{tablegray}{TD3 trained and} \\ \textcolor{tablegray}{evaluated w/o penalty}\end{tabular} & \textcolor{tablegray}{$100$} & \textcolor{tablegray}{$-$} & \textcolor{tablegray}{$-11.07 \pm 0.59$} & \textcolor{tablegray}{$69.2 \pm 4.1$} \\[2mm] \hdashline \\[-2mm]
        POLICEd (ours) & $93.4$ & $\mathbf{93.4}$ & $\mathbf{-16.22 \pm 0.68}$ & $\mathbf{100 \pm 0.0}$ \\[2mm]
        TD3 & $75.8$ & $12.0$ & $-45.20 \pm 3.23$ & $28.4 \pm 3.9$ \\[2mm]
        CPO & $2.0$ & $2.0$ & $-96.71 \pm 3.45$ & $89.9 \pm 2.7$ \\[2mm]
        PPO-Barrier & $\mathbf{100}$ & $86.2$ & $-41.26 \pm -2.30$ & $86.2 \pm 3.0$  \\[1mm] \hdashline[0.9pt/3pt] \\[-3mm]
        \begin{tabular}{@{}c@{}} POLICEd trained \\ w/o penalty \end{tabular} & $48.0$ & $41.6$ & $-70.09 \pm 1.22$ & $41.6 \pm 4.3$ \\[3mm]
        TD3 trained w/o penalty & $99.8$ & $48.8$ & $-45.69 \pm 16.61$ & $53.4 \pm 4.4$ \\ \bottomrule
    \end{tabular}
    \caption{Metrics comparison for different methods based on a 500 episode deployment with the fully-trained policies on the safe arm task illustrated in Figure~\ref{fig: safe KUKA arms}.
    The top row (\textcolor{tablegray}{gray}) provides an upper bound on the completion rate and maximal reward achievable as TD3 is evaluated without penalties for constraint violation (i.e. without $R_{\text{unsf}}$).
    We compare our POLICEd method against the soft-constraint baseline CPO~\citep{constrained_policy_optimization} and the learned safety certificate baseline PPO-Barrier~\cite{yang2023model}. We also report the metrics for TD3 trained with and without the penalty as well as our POLICEd method trained without penalty as part of our ablation study.
    The bold numbers denote the highest values achieved when constraint violations are appropriately penalized.
	The completion task only assess whether the target is eventually reached, even if the constraint is not properly respected.
	For all metrics higher is better.}
    \label{tab: comparison}
\end{table*}

While we believe ours is the first paper to provably enforce hard constraints with black-box environments, safe RL has produced remarkable works in soft constraints and learned safety certificate methods. As such, we also compare our approach with the soft constraint method Constrained Policy Optimization (CPO)~\citep{constrained_policy_optimization}, as well as the learned control barrier function approach PPO-Barrier of \citep{yang2023model}.

As seen in Table~\ref{tab: comparison}, our POLICEd policy is the only algorithm to \emph{guarantee} constraint satisfaction. The soft constraint CPO \citep{constrained_policy_optimization} and the learned safety certificate PPO-Barrier \citep{yang2023model} baselines provide better constraint adherence than standard RL algorithms like TD3, but still fall far short of guaranteed satisfaction. In comparison, our POLICEd approach has the highest task completion percentage without violations by a $40\%$ margin, while also being the highest reward earning policy by nearly a margin of 3 times.

Through our ablation study, we confirm that our POLICEd approach is necessary for constraint satisfaction, as seen by the poor average constraint satisfaction by the TD3 and TD3 trained without penalty policies. Furthermore, it follows intuitively that the constraint violation penalties guide the policy to avoid the region, a concept extensively studied in soft constraint works~\citep{constrained_policy_optimization, review_safe_RL}. As expected, training without penalizing constraint violations is vastly detrimental to the performance of both the TD3 and POLICEd policies. In the POLICEd case, the reward shaping is necessary for the policy to appropriately tune the affine region to avoid the constraint.
We can now answer questions \textbf{Q1}, \textbf{Q2}, and \textbf{Q3} as the POLICEd policy achieves highest average reward while \emph{guaranteeing constraint satisfaction} even on this relatively high-dimensional, high-fidelity robotic arm environment.

We additionally observed that our POLICEd approach exhibited significantly greater sample efficiency compared to our baseline methods. The POLICEd policy converged within 4000 episodes, each consisting of 100 steps. In contrast, CPO frequently failed to converge even after 20,000 episodes of the same length. While PPO-Barrier achieved convergence within 200 iterations, these iterations encompassed numerous episodes of uncapped length, resulting in nearly double the number of environment samples required compared to POLICEd. For additional details see Appendix~\ref{sec: arm lit comparison}.

\section{Conclusion}\label{sec: conclusion}
\noindent \textbf{Summary.} We proposed \emph{POLICEd RL}, a novel algorithm explicitly designed to enforce hard safety constraints for a black-box robot in closed loop with a RL policy.
Our key insight was to build a repulsive buffer around the unsafe area with a locally affine learned policy to guarantee that trajectories never leave the safe set. Our experiments showed that POLICEd RL can enforce hard constraints in high-dimensional, high-fidelity robotics tasks while significantly outperforming existing methods.

\noindent\textbf{Limitations.} With Proposition~\ref{prop: equivalence condition}, we can verify whether there exists a safe POLICEd policy. However, as in standard RL, there are no guarantees that the training will converge to this safe policy. Moreover, in high dimensional environments, the exponential number of vertices of buffer region $\mathcal{B}$ $\big( 2^n$ for a box in dimension $n\big)$ will become computationally prohibitive. However, once trained, a POLICEd DNN is just as fast as a standard unconstrained DNN~\citep{police}.
Another difficulty arising in high-dimensional black-box environments is the computation of $\varepsilon$ and $r$ following~\eqref{eq: approximation} and \eqref{eq: buffer width} respectively. Indeed, these computations require sampling enough transition triplets to bound the environment dynamics. 
Finally, we would like to point out that if the estimate of the approximation measure $\varepsilon$ is too low, then the safety guarantees of Theorem~\ref{thm: admissible trajectories} and Corollary~\ref{cor: discrete repulsion} will not hold. We would like to investigate how an upper bound of $\varepsilon$ can be estimated for general high-dimensional settings.

\noindent\textbf{Future Directions.}
We are excited by our findings and believe our method is only the first step towards enforcing hard constraints on RL policy. For future work, we plan to investigate the case of higher relative-degree constraints by taking ideas from the works that extended CBFs to higher-relative degrees~\citep{ma2021model, Exponential_CBF, HOCBF, BarrierNet}. We would like to further consider enforcing multiple constraints simultaneously. This extension would require minimal changes to our theory but would mostly involve extending the POLICE algorithm \citep{police} to enforce several affine regions instead of just one as initially designed.

\section*{Acknowledgments}

This work is supported by the National Science Foundation, under grants ECCS-2145134, CAREER Award, CNS-2423130, and CCF-2423131.

\bibliographystyle{plainnat}
\bibliography{references}

\appendix

\section{POLICEd RL training tips}\label{sec: training tips}

During our numerous implementations we learned a few tips to help any POLICEd actor to learn a safe policy faster. However, as in standard RL training, there are no guarantees that training will converge to a feasible policy.

To ensure the admissibility of policy $\mu_\theta$, i.e., $\mu_\theta(s) \in \mathcal{A}$ for all $s \in \mathcal{S}$, the classical approach would be to add a final layer to the neural network constituted of a bounded function like the hyperbolic tangent or sigmoid. However, these functions are not piecewise affine, which would invalidate the assumptions of \citep{police}. A clipping of the output would conserve the piecewise affine character of $\mu_\theta$ but would modify the partition $\mathcal{R}$ of \eqref{eq: CPA}, hence preventing the POLICE algorithm to guarantee the affine character of $\mu_\theta$ on buffer $\mathcal{B}$.
Instead, we addressed this issue by adding a large penalty to the reward function when $\mu_\theta(s) \notin \mathcal{A}$. This approach has been very successful in all our experiments.

Another helpful trick to improve the training of a POLICEd network comes from manual curriculum learning. We start the training with an affine buffer $\mathcal{B}$ of size zero. Once a reward threshold is met, we iteratively increase the size of the buffer until it reaches the specified size. This minimizes the impact of POLICE in slowing the learning process. 

Similarly, we noticed that resetting the initial state of some trajectories inside the buffer helped the POLICEd policy learn repulsion condition~\eqref{eq: repulsion} or \eqref{eq: discrete repulsion}. We would typically reset 1 in 10 initial states inside the buffer during training. 

It is also useful to note that for an actor-critic algorithm, only the actor DNN is POLICEd, and the critic is not modified.

We tested several different approaches for the environment behavior with respect to constraint violations.
\begin{itemize}
    \item \emph{Penalized Constraints}: when the agent violates the constraint, it is allowed to keep going but incurs a penalty.
    \item \emph{Terminal Constraints}: when the agent violates the constraint, the episode ends with a penalty.
    \item \emph{Bouncing Constraints}: when the agent tries to step into the constraint, it incurs a penalty and its state remains at its last constraint-abiding value.
\end{itemize}
The inverted pendulum environment of Section~\ref{subsec: pendulum} implements terminal constraints, while the linear environment of Section~\ref{sec: implementation} and the robotic arm environment of Section~\ref{subsec: arm} rely on bouncing constraints.

\begin{algorithm}[htbp!]
\caption{POLICEd RL training process} \label{alg: training}
\begin{algorithmic}[1]
\Require Environment $E$~\eqref{eq: discrete observations}, constraint~\eqref{eq: constraint}, reward threshold $r_{T}$
\State Let $\mathcal{R}_b$ denote the replay buffer, $\mu_{\theta}$ our policy, $\mathcal{B}$ our affine buffer, and $\mathcal{U}$ the uniform sampling function.
\State \Comment{Main training loop}
\While{reward < $r_{T}$ \textbf{and} \eqref{eq: discrete repulsion} not satisfied on $\mathcal{B}$}
\While{reward < $r_{T}$} \Comment{Standard RL training}
\State Reset $s(0) = 0$ or $s(0) \sim \mathcal{U}(\mathcal{B})$
\While{not $done$}
\State Take action $a(t) = \mu_\theta( s(t) )$
\State Get reward $r(t) = R(s(t), a(t))$
\State Get ($s(t+\delta t)$, $respect$, $done$) from $E$
\State Add $\big( s(t), a(t), s(t+\delta t), r(t), done \big)$ to $\mathcal{R}_b$
\EndWhile
\State Update policy $\mu_\theta$ with $\mathcal{R}_b$ \Comment{Note: This often reduces the constraint satisfaction of the policy}
\EndWhile

\State \Comment{Constraint Training Loop}
\While{\eqref{eq: discrete repulsion} not satisfied on $\mathcal{B}$} 
\State Reset $s(0) \sim \mathcal{U}(\mathcal{V}(\mathcal{B}))$\Comment{Reset the initial state only on the buffer vertices}
\State Take action $a(0) = \mu_\theta( s(0) )$
\State Get reward $r(t) = R(s(0), a(0))$
\State Get ($s(\delta t)$, $respect$, $done$) from $E$
\If{not $respect$}
    \State Add $\big( s(0), a(0), s(\delta t), r(0), done \big)$ to $\mathcal{R}_b$
\EndIf
\State Update policy $\mu_\theta$ with $\mathcal{R}_b$ \Comment{Note: This often reduces the reward earned by the policy}
\EndWhile

\EndWhile %main loop

\Ensure Trajectories $\tau_d^{\mathcal{S}}(s_0, \mu_\theta)$ of \eqref{eq: discrete traj} starting from safe state $s_0 \in \mathcal{S}_s$ do not leave safe set $\mathcal{S}_s$ 
\end{algorithmic}
\end{algorithm}

\section{Implementation Details}\label{apx: details}

\subsection{Safe Inverted Pendulum}

In the inverted pendulum experiment, we choose to limit the buffer to $x \in [-0.9, 0.9]$, whereas the cart position $x$ can in fact reach $-1$ and $1$. However, when reaching these extremal positions the cart bounces back which provokes a severe discontinuity in the velocities. Our framework can handle discontinuities through the $\varepsilon$ term of~\eqref{eq: approximation}
Moreover, a safe policy must be able to stabilize the inverted pendulum without pushing the cart to its extremal positions. For all these reasons, we decided to stop our buffer short of $x = \pm 1$.

\subsection{Safe Robotic Arm}

At each timestep $k$, the actions $a_k \in \mathbb{R}^7$ outputted by the policy are promoted to only produce small joint changes in each step. For the KUKA Arm, the maximum joint angles are 
$$[\pm2.97, \pm2.09, \pm2.97, \pm2.09, \pm2.97, \pm2.09, \pm3.05],$$
and the agent incurs a penalty of 3 for choosing an action which would go outside of this range.

The target is always placed at X-Y-Z coordinates $[0.5, 0.5, 0.5]$, the constraint is a rectangular prism region placed above the target and centered at $[0.5, 0.5, 0.60]$ with side-lengths $[0.16, 0.30, 0.06]$, respectively. We choose this constraint because it prevents the robotic arm from easily reaching the target and from fully minimizing the reward. Indeed, the optimal policy is modified by the existence of the constraint for a large set of initial states $s(0)$. The buffer $\mathcal{B}$ is larger than the constraint and surrounds it on all sides. The buffer is centered at $[0.5, 0.5, 0.62]$ and is also a rectangular prism with sizes $[0.18, 0.32, 0.10]$. These placements are all illustrated in Figure~\ref{fig: safe KUKA arms}. These choice of side-lengths were calculated based on the equations presented in Section~\ref{sec: framework}.

Additionally, during training, we also prevented the agent from taking actions that would violate the constraint during training, which involves both assigning a penalty and leaving the state unchanged. This choice was made to ensure that the episodes that involved violation did not end too quickly, as longer episodes that did not violate the constraint might take more steps and incur more reward penalties. When evaluating the baselines, both training with violations allowed and training without were tested, and the case with better performance was reported.

\subsubsection{Comparison with the literature}\label{sec: arm lit comparison}

While building Table~\ref{tab: comparison} we noticed several interesting facts about other approaches. We found that the CPO~\citep{constrained_policy_optimization} method becomes increasingly incapable of approaching the target. Additionally, given that the state space is very large, it can often experience new initial states during deployment, causing it to still violate the constraint in some cases.

\begin{figure}[htb!]
    \centering
    \subfloat[Averaged log-reward]{\includegraphics[scale=0.53]{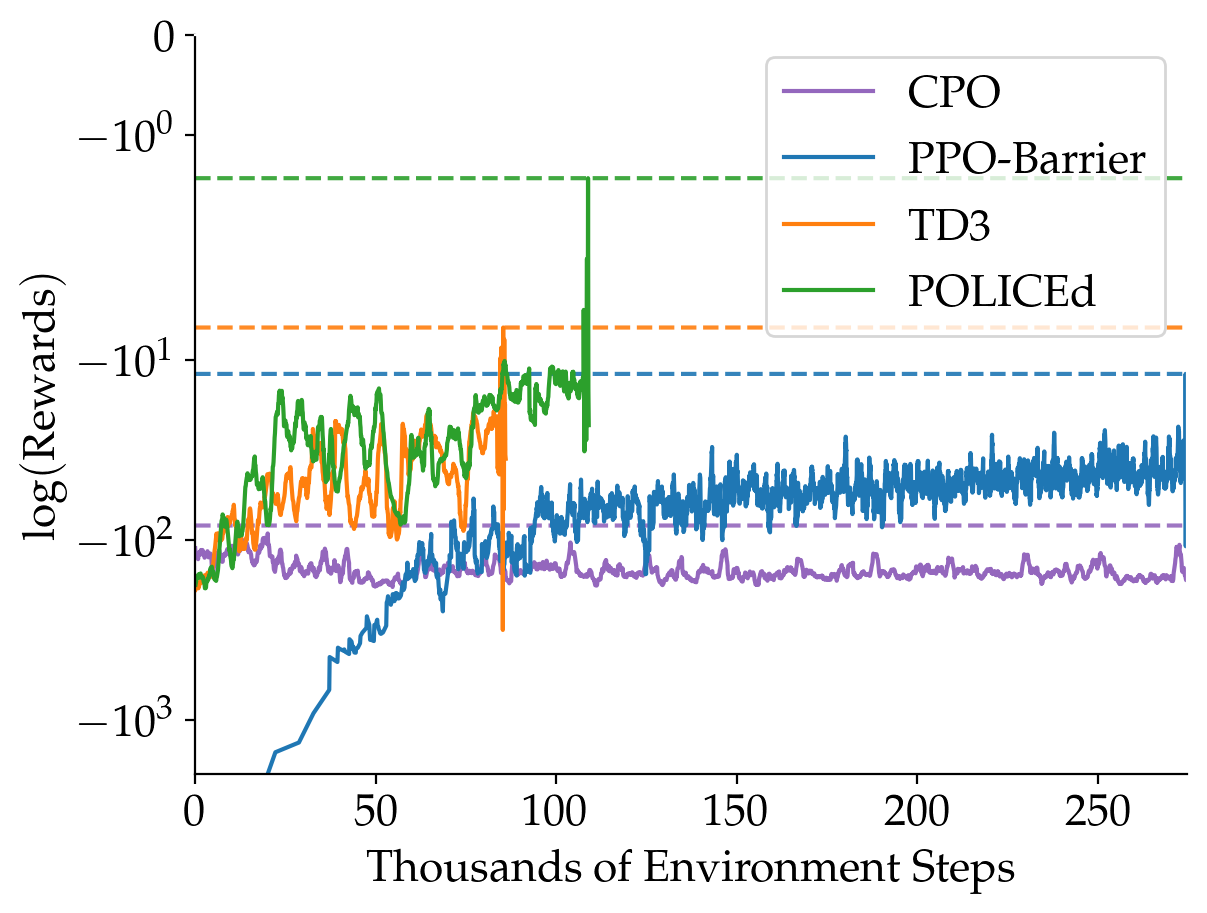}}\hfill
    \subfloat[Averaged percentage of constraint satisfaction]{\includegraphics[scale=0.53]{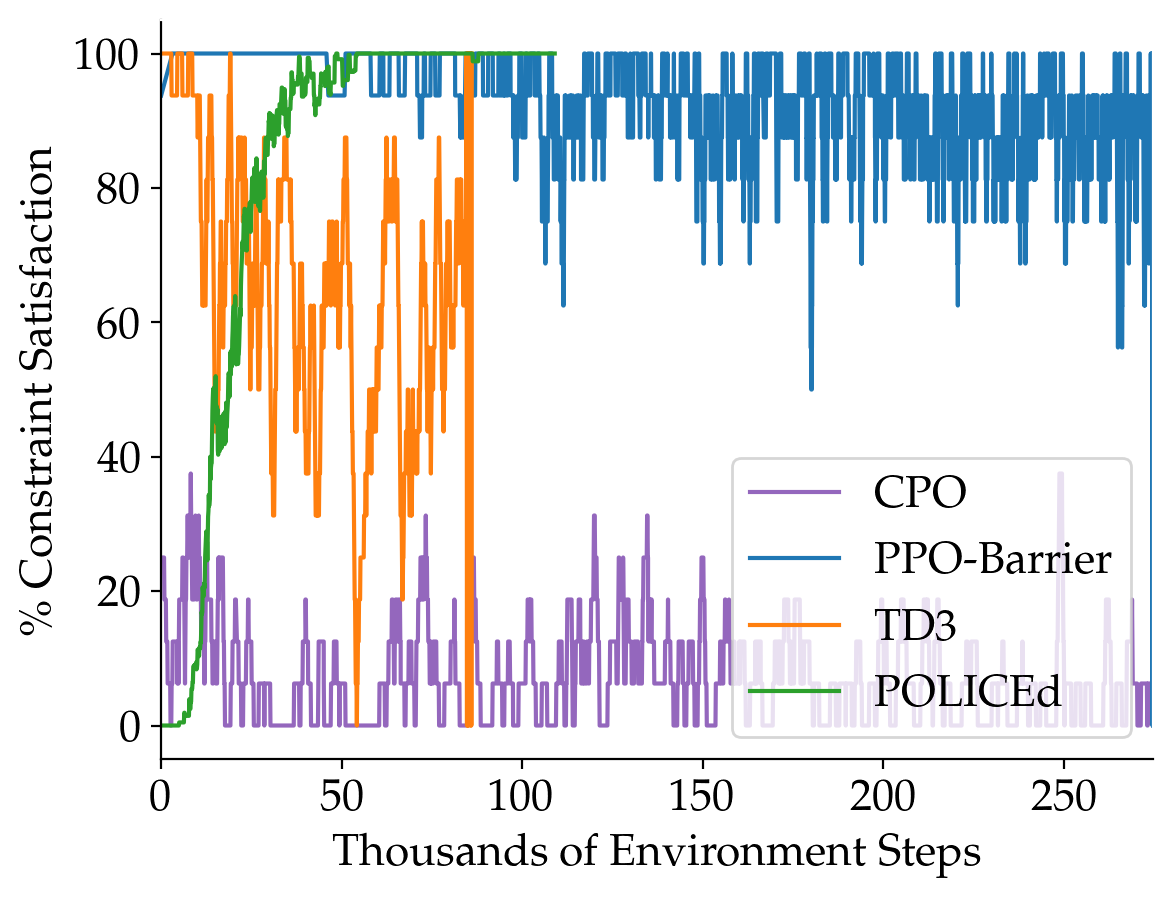}}
    \caption{Reward and constraint adherence curves for the safe arm scenario. For the average reward curve, the solid lines correspond to the reward per episode averaged and $log(\cdot)$ over 5 training cycles of each method. The dashed lines are the highest average rewards achieved by each method. Meanwhile for the constraint satisfaction, the constraint respect per episode was averaged over 5 training cycles of each method. Note that the x-axis on both curves is in thousands of environment samples, and each curve ends when it was considered converged. Note that CPO was trained for an average of 4 million environment steps and has been cut off, but the reported maximum reward is over the entire dataset. 95\% CI bounds were omitted for clarity, but can be provided on request.}
    \label{fig: Arm Training Curves}
\end{figure}

Meanwhile, the PPO-Barrier~\citep{yang2023model} approach often gathers many thousands of episodes and environment samples in its initial "iterations", allowing it to appear to train quickly (around 250,000 environment steps for convergence). In comparison, our POLICEd approach primarily converges in around 150,000 environment samples. We further showcase these findings in the training curves showcased in Figure~\ref{fig: Arm Training Curves}.

\end{document}